\documentclass[11pt]{article}

\usepackage[preprint]{acl}

\usepackage{times}
\usepackage{latexsym}

\usepackage[T1]{fontenc}

\usepackage[utf8]{inputenc}

\usepackage{microtype}

\usepackage{inconsolata}

\usepackage{graphicx}
\usepackage{booktabs}
\usepackage{multirow}
\usepackage{amsmath}
\usepackage{amssymb}

\graphicspath{{../figures/}{figures/}{./}}

\title{Axis-LoRA: Probing and Adapting Embedding Models for Personal-Memory Retrieval}

\author{
Dohoon Kim$^{1}$ \quad
    Junho Park$^{1}$ \quad
    Hyunwoong Bae$^{2}$ \quad
    Seyoon Kim$^{2,3}$ \quad
    Taesup Moon$^{1,2,4}$\thanks{Corresponding author}
    \\
    $^{1}$ Department of Electrical and Computer Engineering, Seoul National University \\
    $^{2}$ Interdisciplinary Program in Artificial Intelligence, Seoul National University \\
    $^{3}$ Mobile eXperience Business, Samsung Electronics \\
    $^{4}$ ASRI/INMC/AIIS, Seoul National University \\
    {\tt\small \{dohoon.kim, bryan123, hwbae0326, sykimy, tsmoon\}@snu.ac.kr }
}

\begin{document}
\maketitle
\begin{abstract}
As LLMs are increasingly asked questions about a user's personal experiences, retrieval must distinguish memories by time, place, person, and event. Current personal-memory pipelines rely on frozen dense embedding models to embed memories for retrieval, but we find that these models do not capture the four axes well. Swapping a memory's date or place barely shifts its embedding, and a trained projection on top of the frozen embedding does not recover the missing signal. The problem lives inside the model, where each axis is encoded by a distinct attention-head pattern. Guided by these findings, we propose Axis-LoRA, a per-axis adapter that places four LoRA branches inside the embedding model and combines them via a per-query gate. Across three standard embedding models evaluated on LongMemEval, ATM-Bench, and LoCoMo, Axis-LoRA improves 4-benchmark mean R@10 by $+1.28$ to $+1.56$~pp while training only a small fraction of the base parameters on a small axis-counterfactual training set. Code is available at \url{https://github.com/dohoonkim-ai/axis-lora}.
\end{abstract}

\section{Introduction}
\label{sec:intro}

\begin{figure*}[t!]
  \centering
  \includegraphics[width=1.0\textwidth]{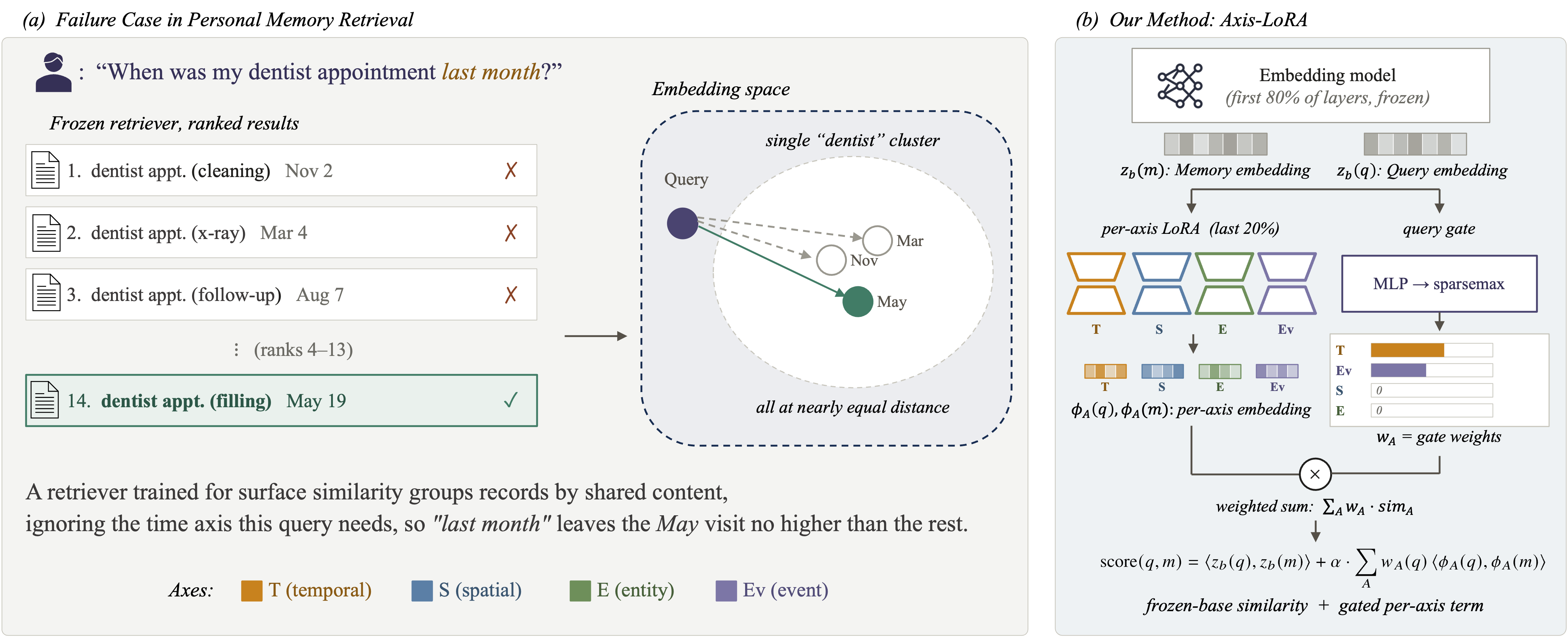}
  \caption{Overview. \textbf{(a) Failure case in personal-memory retrieval.} A frozen retriever groups same-content memories into a single embedding cluster, where shared content tokens place candidates at near-equal cosine distance. An axis cue in the query (here, the time axis) does not separate the gold from content-similar distractors that mismatch on the axis, so the gold falls below them in the ranked list. \textbf{(b) Axis-LoRA.} Four axis-conditional LoRA adapters (T/S/E/Ev) attach to the last $20\%$ of layers of the frozen embedding model and produce per-axis representations $\phi_A(q), \phi_A(m)$. A per-query gate (MLP $\to$ sparsemax) outputs routing weights $w_A$ with exact zeros on inactive axes. The final score is the frozen-base similarity plus a gated per-axis term.}
  \label{fig:teaser}
\end{figure*}

Large language models (LLMs) are increasingly expected to answer questions about a user's personal experiences (e.g., \emph{``What did I discuss with Marco at Tuesday's meeting?''} or \emph{``Where did I park at the Boston conference?''}), not just open-domain factual queries \citep{packer2023memgpt, chhikara2024mem0, gutierrez2024hipporag}. Several benchmarks have been introduced to evaluate this \emph{personalized memory QA} ability: LongMemEval \citep{wu2024longmemeval} and LoCoMo \citep{maharana2024locomo} target long-term multi-session conversations, with LoCoMo spanning multiple months, while ATM-Bench \citep{mei2026atmbench} extends memory to multi-source personal data (images, videos, emails).

Retrieval is central to personalized memory QA: given a finite LLM context window, surfacing the query-relevant memories from a user's long memory corpus largely determines answer quality \citep{lewis2020rag}. To support retrieval over long personal memory corpora, a line of recent work develops memory-management pipelines with hierarchical and long-term storage \citep{chhikara2024mem0, kang2025memoryos, li2025memos, hu2026evermemos}, yet uniformly relies on a general-purpose embedding model \citep{karpukhin2020dpr, izacard2022contriever} for the retrieval itself.

Personal memory, however, differs in structure from general retrieval corpora: each memory is naturally indexed along a small set of categorical axes, of which we consider four: TEMPORAL (timestamps, dates), SPATIAL (places, locations), ENTITY (persons, objects), and EVENT (actions, event types). A useful personalized memory retriever must discriminate along each of these axes. General embedding model training \citep{karpukhin2020dpr, izacard2022contriever, gao2021simcse, wang2022e5}, however, optimizes encoders for topic and content overlap, so it under-trains them on the per-memory attributes that distinguish otherwise topically similar episodic records.

We confirm this gap on three standard embedding models. Swapping a memory's axis value (e.g., replacing \emph{Tuesday} with \emph{Friday}) barely changes its embedding, and the retriever therefore fails to rank the right memory near the top when a query targets that axis. To understand where the signal is missing, we run two diagnostic probes \citep{goldfarb-tarrant2024multicontrievers}. A learned projection trained on top of the frozen embedding cannot recover it, but a per-attention-head analysis shows each axis concentrated in a \emph{distinct} attention-head pattern. The signal lives inside the model, not at its output.

Guided by these findings, we propose a simple per-axis adapter, \textbf{Axis-LoRA} (Figure~\ref{fig:teaser}b): four axis-conditional LoRA \citep{hu2022lora} branches inside the frozen embedding model, gated per query. We train Axis-LoRA on AxisCF, a synthetic axis-counterfactual triplet set, and evaluate its retrieval R@10 on three dense embedding models (MiniLM-L6-v2, bge-base-en-v1.5, Qwen3-Embedding-0.6B) over LongMemEval, ATM-Bench, and LoCoMo, consistently improving the 4-benchmark mean R@10 on all three embedding models.

Our contributions are summarized as follows:
\begin{itemize}
  \item We show that frozen dense encoders exhibit \emph{axis-asymmetric under-exposure} of personal-memory attributes (each axis is under-exposed to a different degree), evident both in axis-counterfactual cosine distance and in axis-anchored R@10.
  \item Two diagnostic probes localize the cause inside the encoder: an external projection over the frozen output cannot recover the gap, and each attribute draws from a distinct attention-head footprint, which naturally motivates an inside-encoder, axis-specific fix.
  \item We propose Axis-LoRA, a simple per-axis adapter with four axis-conditional LoRA branches and a per-query gate, trained on AxisCF, our synthetic axis-counterfactual triplet set. On LongMemEval, ATM-Bench, and LoCoMo, Axis-LoRA improves mean R@10 across the three encoders by $+1.28$ to $+1.56$~pp, with the largest per-bench gains on the multi-evidence ATM-Bench Hard split ($+2.4$ to $+3.8$~pp across the three encoders).
\end{itemize}

\section{Phenomenon: Axis-Asymmetric Under-Exposure}
\label{sec:phenomenon}

Dense embedding encoders expose the four memory axes (TEMPORAL, SPATIAL, ENTITY, EVENT) to different degrees, yet all four sit far below the cosine-distance scale required for retrieval to discriminate candidates. We characterize this \emph{axis-asymmetric under-exposure} at two levels: in the embedding (\S\ref{sec:phenomenon-structural}) and in axis-anchored retrieval (\S\ref{sec:phenomenon-behavioral}).

\paragraph{Setup: Axis-isolated counterfactuals.}
To isolate the effect of each axis, we use \textbf{AxisCF}, a synthetic corpus of axis-isolated counterfactuals. For each memory $M$, AxisCF provides four counterfactuals $\mathrm{CF}_A$, one per axis $A \in \{\textsc{Temp}, \textsc{Spat}, \textsc{Ent}, \textsc{Event}\}$. Each $\mathrm{CF}_A$ differs from $M$ only in tokens of axis $A$, leaving non-axis content unchanged (Table~\ref{tab:cf-example}). Pool composition and axis balancing are detailed in Appendix~\ref{sec:appendix-axiscf-eval-pool}, and corpus generation in Appendix~\ref{sec:appendix-axiscf-generation}.

\begin{table}[t]
  \centering
  \footnotesize
  \setlength{\tabcolsep}{4pt}
  \caption{Axis-isolated counterfactual example. Each $\mathrm{CF}_A$ swaps the value of axis $A$ while keeping the rest of the memory unchanged.}
  \label{tab:cf-example}
  \begin{tabular}{@{}lp{0.76\columnwidth}@{}}
    \toprule
    $M$ & I met Marco at the Boston conference on 2024-06-15 to discuss the rehearsal dinner. \\
    \midrule
    $\mathrm{CF}_{\textsc{Temp}}$  & ``2024-06-15'' $\to$ ``2024-09-20'' \\
    $\mathrm{CF}_{\textsc{Spat}}$  & ``Boston'' $\to$ ``Seattle'' \\
    $\mathrm{CF}_{\textsc{Ent}}$   & ``Marco'' $\to$ ``Sarah'' \\
    $\mathrm{CF}_{\textsc{Event}}$ & ``rehearsal dinner'' $\to$ ``budget meeting'' \\
    \bottomrule
  \end{tabular}
\end{table}

\begin{figure}[t]
  \centering
  \includegraphics[width=\columnwidth]{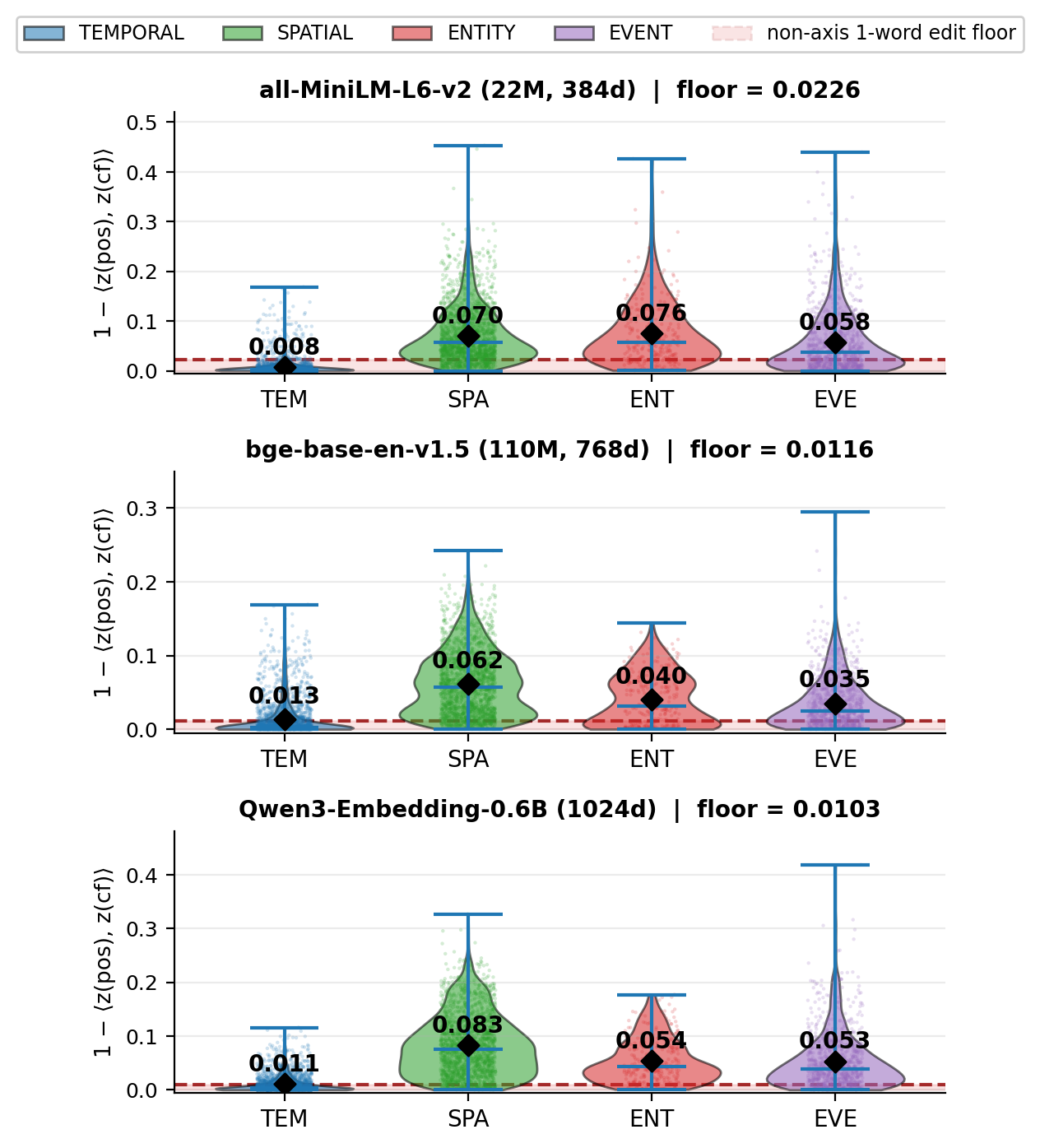}
  \caption{Axis-counterfactual cosine distance per axis across three embedding models. Violins show the distribution over the evaluation pool; diamonds mark the per-axis mean. The red dashed line is the per-model non-axis 1-word edit floor. TEMPORAL sits at this floor on every model, while SPATIAL, ENTITY, and EVENT clear the floor but remain small in absolute magnitude.}
  \label{fig:axis-blindness}
\end{figure}

\subsection{Embedding-level: Axis-swap cosine response}
\label{sec:phenomenon-structural}

Figure~\ref{fig:axis-blindness} plots the per-axis cosine distance $1 - \cos(\mathbf{e}(M), \mathbf{e}(\mathrm{CF}_A))$ on the pooled embeddings across three embedding models: two BERT-style encoder-only models, \texttt{all-MiniLM-L6-v2} \citep{wang2020minilm} ($22$\,M) and \texttt{bge-base-en-v1.5} \citep{xiao2024bge} ($110$\,M), and one causal-attention LLM-based embedding model, \texttt{Qwen3-Embedding-0.6B} \citep{qwen3embedding2025} ($600$\,M). As a reference scale we use a \emph{non-axis 1-word edit floor}: one content word per memory swapped with a random vocabulary word, axis values excluded. We further compare against the retrieval-relevant scale $\mu_{\mathrm{inter}}$, the mean cosine distance between any two memories in the corpus (per-encoder values $0.5$--$0.8$; Table~\ref{tab:axis-scale-graded} and Appendix~\ref{sec:appendix-axis-blindness}). All four axis swaps lie far below this scale.

The ordering is consistent across all three encoders: TEMPORAL is the smallest, with EVENT, SPATIAL, and ENTITY larger but all far below $\mu_{\mathrm{inter}}$. Table~\ref{tab:axis-scale-graded} reports two ratios per cell. Against the non-axis 1-word edit floor, TEMPORAL is at-floor ($0.3$--$1.1\times$, effectively at the noise scale across all three models), while SPATIAL, ENTITY, and EVENT clear the floor by $2.6$--$8.0\times$. Against the retrieval-relevant scale $\mu_{\mathrm{inter}}$, however, all four axes sit at only $1$--$14\%$. The under-exposure pattern holds across all three models. It also holds beyond synthetic AxisCF memories: repeating the measurement on real memories from the three benchmarks ($200$ ATM-Bench, $200$ LoCoMo, $100$ LongMemEval, with axis counterfactuals built by the same pipeline) reproduces the pattern, with TEMPORAL at or below the 1-word edit floor on all three encoders and every axis at $\leq 11\%$ of $\mu_{\mathrm{inter}}$ (Appendix~\ref{sec:appendix-real-memory}).

\begin{table}[t]
  \centering
  \footnotesize
  \setlength{\tabcolsep}{3pt}
  \caption{Per-model, per-axis mean axis-CF cosine distance, expressed against two reference scales. \textbf{Top:} multiple of the non-axis 1-word edit floor $f_e$. \textbf{Bottom:} percentage of the inter-memory distance $\mu_{\mathrm{inter}}$. TEMPORAL is at-floor on every model, and every axis sits at $\leq 14\%$ of $\mu_{\mathrm{inter}}$.}
  \label{tab:axis-scale-graded}
  \resizebox{\columnwidth}{!}{%
  \begin{tabular}{@{}lcccc@{}}
    \toprule
    \textbf{Encoder} & \textbf{TEMP.} & \textbf{SPAT.} & \textbf{ENT.} & \textbf{EVENT} \\
    \midrule
    \multicolumn{5}{l}{\emph{Multiple of non-axis 1-word edit floor $f_e$}} \\
    MiniLM ($f_e\!=\!0.023$)         & $0.34\times$ & $3.12\times$ & $3.35\times$ & $2.55\times$ \\
    bge-base ($f_e\!=\!0.012$)       & $1.12\times$ & $5.29\times$ & $5.78\times$ & $3.01\times$ \\
    Qwen3-Emb-0.6B ($f_e\!=\!0.010$) & $1.04\times$ & $8.02\times$ & $5.23\times$ & $5.13\times$ \\
    \midrule
    \multicolumn{5}{l}{\emph{Percentage of inter-memory distance $\mu_{\mathrm{inter}}$}} \\
    MiniLM ($\mu\!=\!0.78$)          & $1.0\%$ & $9.0\%$  & $9.7\%$  & $7.4\%$ \\
    bge-base ($\mu\!=\!0.47$)        & $2.8\%$ & $13.2\%$ & $14.4\%$ & $7.5\%$ \\
    Qwen3-Emb-0.6B ($\mu\!=\!0.65$)  & $1.6\%$ & $12.7\%$ & $8.3\%$  & $8.1\%$ \\
    \bottomrule
  \end{tabular}}
\end{table}

\subsection{Retrieval-level: Axis-anchored R@10}
\label{sec:phenomenon-behavioral}

We next test whether the embedding-level gap translates to retrieval failure, using an \emph{axis-anchored retrieval} task: each query is a short sentence built from a single axis value of a memory $M$ in the AxisCF evaluation pool. For example, a TEMPORAL query is built from $M$'s date as \emph{``This event happened on 2024-06-15.''}, and an ENTITY query from $M$'s named person as \emph{``This memory involves Marco.''} (per-axis templates in Appendix~\ref{sec:appendix-axiscf-eval-pool}). The retriever ranks all memories in the pool by cosine similarity to the query, with $M$ as the gold target.

Table~\ref{tab:axis-anchored-fair} (\textbf{raw} columns) reports raw cosine R@10. Across the $12$ (encoder, axis) cells, R@10 ranges from $0.039$ (Qwen3 TEMPORAL) to $0.312$ (Qwen3 EVENT), with TEMPORAL the lowest on every encoder ($0.039$--$0.072$). Together with the embedding-level finding of \S\ref{sec:phenomenon-structural}, this confirms that axis information is under-exposed in frozen-embedding cosine retrieval.

\begin{table}[t]
  \centering
  \footnotesize
  \setlength{\tabcolsep}{3pt}
  \caption{Axis-anchored retrieval R@10 on the evaluation set: raw cosine vs.\ projected (per-axis MLP trained with InfoNCE), with $\Delta = $ projected $-$ raw. TEMPORAL gets the largest projection lift, while non-TEMPORAL cells stay flat or decrease.}
  \label{tab:axis-anchored-fair}
  \resizebox{\columnwidth}{!}{%
  \begin{tabular}{@{}lccc|ccc|ccc@{}}
    \toprule
     & \multicolumn{3}{c|}{\textbf{MiniLM}} & \multicolumn{3}{c|}{\textbf{bge-base}} & \multicolumn{3}{c}{\textbf{Qwen3-0.6B}} \\
    \textbf{Axis} & raw & proj & $\Delta$ & raw & proj & $\Delta$ & raw & proj & $\Delta$ \\
    \midrule
    TEMP.  & $0.061$ & $0.150$ & $+8.9$  & $0.072$ & $0.167$ & $+9.5$  & $0.039$ & $0.183$ & $+14.4$ \\
    SPAT.  & $0.248$ & $0.251$ & $+0.3$  & $0.305$ & $0.275$ & $\mathbf{-3.0}$ & $0.231$ & $0.233$ & $+0.2$ \\
    ENT.   & $0.163$ & $0.224$ & $+6.1$  & $0.154$ & $0.179$ & $+2.5$  & $0.152$ & $0.197$ & $+4.5$ \\
    EVENT  & $0.254$ & $0.260$ & $+0.6$  & $0.295$ & $0.266$ & $\mathbf{-2.9}$ & $0.312$ & $0.231$ & $\mathbf{-8.1}$ \\
    \bottomrule
  \end{tabular}}
\end{table}

\section{Diagnosis: Why Axis Signals Are Under-Exposed}
\label{sec:diagnosis}

The under-exposure shown in \S\ref{sec:phenomenon} raises two questions. \emph{(i)} Can the missing axis discrimination be recovered from the frozen output by a learned head (i.e., does the encoder already surface it but in a geometry that cosine retrieval cannot exploit)? \emph{(ii)} If not, where inside the embedding model does each axis signal live, and is the internal footprint shared across the four axes or axis-specific? Two diagnostic probes follow: a \emph{projection probe} (\S\ref{sec:diagnosis-p1}) tests (i), and a \emph{head-spectrum probe} (\S\ref{sec:diagnosis-p2}) tests (ii).

\subsection{Projection probe: External projection cannot close the R@10 gap}
\label{sec:diagnosis-p1}

For TEMPORAL, which sits at the non-axis edit floor, is the axis signal \emph{erased} during encoding, or merely \emph{latent-compressed} (preserved in the frozen output and recoverable by a learned projection)? For SPATIAL, ENTITY, and EVENT (raw cosine R@10 $0.17$--$0.33$), can an external projection improve R@10, or is this the ceiling an output-side head can reach?

\paragraph{Setup.} We train a strong external projection: a per-axis 2-layer MLP head $\phi_A$ over the frozen memory embeddings, optimized with InfoNCE. Positives are $(q_A, \phi_A(m))$; hard negatives are the axis-$A$ counterfactuals $\mathrm{CF}_A$ of $m$ from \S\ref{sec:phenomenon}. We evaluate on the same setup as \S\ref{sec:phenomenon-behavioral}. The setup isolates the question: what can we achieve with the \emph{frozen latent plus axis-CF training signal alone}, without modifying the encoder? Full training and evaluation details, including two simpler projection variants we considered, are in Appendix~\ref{sec:appendix-projection-probe}.

\paragraph{Two findings.}
\emph{(i) The axis signal is not fully erased, but recovery is small.} The projection lifts TEMPORAL R@10 by $+8.9$ to $+14.4$~pp across all three encoders (Table~\ref{tab:axis-anchored-fair}), confirming that some axis information survives in the latent and is reachable by an output-side head. The absolute R@10 ceiling, however, stays at $0.15$--$0.18$, far short of useful retrieval. An external head does not close the gap.

\emph{(ii) Projection effects vary across axes.} ENTITY lifts consistently ($+2.5$ to $+6.1$~pp), while SPATIAL and EVENT are mixed, with per-cell deltas ranging from small positive lifts to drops of up to $-8.1$~pp. An output-side projection therefore does not provide a uniform fix across axes.

\paragraph{Implication.}
Axis information is partially recoverable, but a strong external head cannot bridge the R@10 gap and does not uniformly help across non-TEMPORAL axes. This rules out an output-side fix and points to modifying the embedding model itself, changing how axis information is encoded internally rather than how the frozen output is post-processed.

\subsection{Head-spectrum probe: The attention-head spectrum is axis-asymmetric}
\label{sec:diagnosis-p2}

The projection probe established that axis information lives \emph{inside} the embedding model. To find where each axis signal lives, we run a per-(layer, head) analysis: for each (layer $\ell$, head $h$) and axis $A$, we measure the change in the pooled per-head output when the input is perturbed by a single-edit $\mathrm{CF}_A$ that alters exactly one token (Appendix~\ref{sec:appendix-head-spectrum-robustness}):
\begin{equation*}
  \mathrm{resp}[\ell, h, A]
  = \frac{\| \mathrm{head}_{\ell, h}(M) - \mathrm{head}_{\ell, h}(\mathrm{CF}_A) \|_2}
         {\| \mathrm{head}_{\ell, h}(M) \|_2},
\end{equation*}
averaged over many $(M, \mathrm{CF}_A)$ pairs, where $\mathrm{head}_{\ell, h}(M) \in \mathbb{R}^{d_h}$ is the per-head output after token-pooling (last-token for Qwen3-Embedding-0.6B, mean-pool for MiniLM and bge-base).

\begin{figure}[t]
  \centering
  \includegraphics[width=\columnwidth]{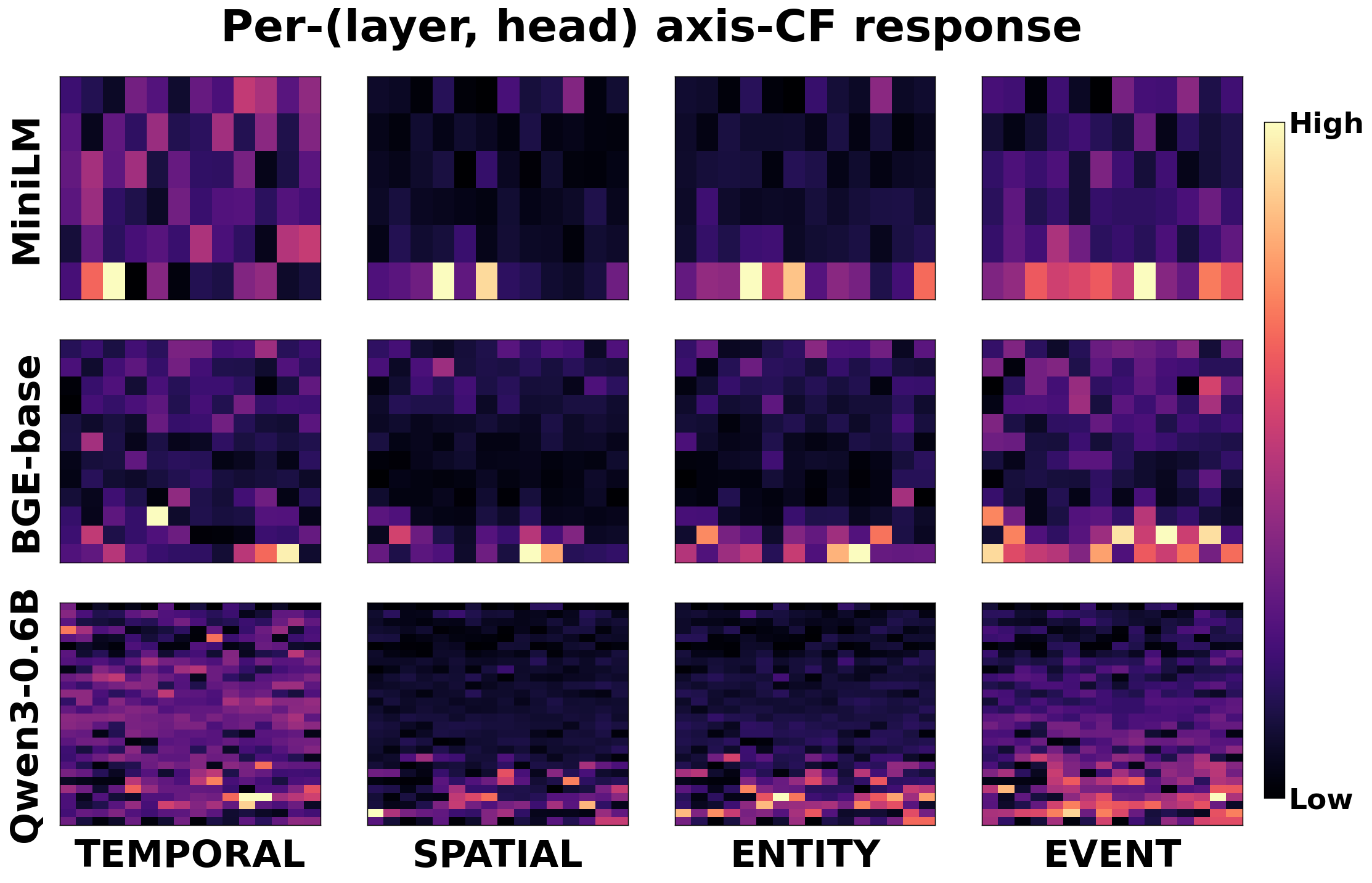}
  \caption{Head-spectrum probe: per-(layer, head) axis response for three embedding models (rows) $\times$ four axes (columns). Within each panel, rows index transformer layers (top: input side, bottom: output side) and columns index attention heads. Brightness shows how much a head's pooled output changes under single-edit axis perturbation, normalized per panel to a shared Low--High scale. On every model and axis, bright cells cluster in the later layers.}
  \label{fig:p2-heatmap}
\end{figure}

\begin{table}[t]
  \centering
  \footnotesize
  \setlength{\tabcolsep}{4pt}
  \caption{Fraction of attention heads needed to cover $50\%$ of per-axis single-edit response. A small fraction means the axis is concentrated in few heads, a large fraction that it is spread across many; the largest per model (most diffuse axis) is bolded. Within each model, TEMPORAL is the most diffuse, EVENT next, and SPATIAL and ENTITY are the most concentrated.}
  \label{tab:p2-50pct-heads}
  \resizebox{0.8\columnwidth}{!}{%
  \begin{tabular}{@{}lcccc@{}}
    \toprule
    \textbf{Model} & \textbf{TEMP.} & \textbf{EVENT} & \textbf{SPAT.} & \textbf{ENT.} \\
    \midrule
    MiniLM       & \textbf{41.7\%} & 36.1\% & 29.2\% & 26.4\% \\
    bge-base     & \textbf{39.6\%} & 35.4\% & 29.2\% & 30.6\% \\
    Qwen3-0.6B   & \textbf{32.1\%} & 25.4\% & 19.6\% & 18.5\% \\
    \bottomrule
  \end{tabular}}
\end{table}

\paragraph{Axis response concentrates in late layers.}
Figure~\ref{fig:p2-heatmap} shows the per-(layer, head) response across the three models, with brighter cells indicating heads whose pooled output changes more under axis perturbation. We find a consistent pattern: on every model and every axis, the bright cells cluster in the later layers, indicating that axis-sensitive computation is concentrated in the late layers of the embedding model.

\paragraph{Each Axis draws distinct head footprint.}
To quantify diffuseness, for each axis we report the percentage of top-ranked heads that together account for $50\%$ of axis-$A$ response, as shown in Table~\ref{tab:p2-50pct-heads}. A small percentage means the axis is concentrated in few heads, a large percentage that it is spread across many. Within each model, TEMPORAL is the most diffusely encoded, EVENT next, and SPATIAL and ENTITY are the most concentrated. This ordering is consistent across all three models. The same pattern is visible in Figure~\ref{fig:p2-heatmap}: the TEMPORAL and EVENT panels have bright cells spread across many heads, while the SPATIAL and ENTITY panels have bright cells concentrated in fewer positions. These results establish that each axis has its own response pattern across the heads of the embedding model.

\section{Method: Axis-LoRA}
\label{sec:method}

The diagnosis in \S\ref{sec:diagnosis} naturally points to a simple per-axis adapter. We propose Axis-LoRA, whose three core design choices follow directly from the findings. The projection probe (\S\ref{sec:diagnosis-p1}) shows that an external projection over the frozen latent cannot bridge the R@10 gap, so Axis-LoRA reshapes the embedding model internally rather than projecting its output. The head-spectrum probe (\S\ref{sec:diagnosis-p2}) shows each axis drawing from a distinct head footprint, motivating four independent LoRA adapters. The same probe also shows axis-sensitive computation concentrated in the late layers of the embedding model, so we restrict LoRA to its late attention layers.

\subsection{Architecture}
\label{sec:method-architecture}

Axis-LoRA scores a (query, memory) pair by adding a gated sum of per-axis similarities to the frozen base similarity. Each per-axis similarity is computed by a per-axis LoRA adapter inside the same embedding model, and the gate decides per query how much weight each axis carries. The retrieval score for query $q$ and memory $m$ is:
\begin{equation*}
  \begin{aligned}
    \mathrm{score}(q, m) =\ & \langle \tilde{\mathbf{z}}_b(q),\, \tilde{\mathbf{z}}_b(m) \rangle \\
    {}+{} & \alpha \sum_{A \in \mathcal{A}} w_A(q) \cdot \langle \tilde{\boldsymbol{\phi}}_A(q),\, \tilde{\boldsymbol{\phi}}_A(m) \rangle,
  \end{aligned}
\end{equation*}
where $\mathbf{z}_b(\cdot)$ is the frozen base embedding model's pooled output, $\boldsymbol{\phi}_A(\cdot)$ is the same embedding model modified by the LoRA adapter for axis $A$, $\tilde{\cdot}$ denotes $\ell_2$-normalization, $\mathcal{A} = \{\textsc{Temp}, \textsc{Spat}, \textsc{Ent}, \textsc{Event}\}$, $w_A(q)$ is the per-query gate weight on axis $A$, and $\alpha \in [0, 1]$ is the axis-term mixing weight.

\paragraph{Per-axis LoRA adapters.}
Each adapter attaches to the \texttt{q\_proj} and \texttt{v\_proj} projections of the last $20\%$ of attention layers, with LoRA rank $r = 16$. The four adapters do not share parameters, so each axis can learn its own reshape direction without cross-axis interference.

\paragraph{Query-conditioned axis gating.}
A query depends on only a few axes, so the gate is designed to select them and amplify their signal: $w_A(q) = \mathrm{sparsemax}\!\big(\mathrm{MLP}_g(\mathbf{z}_b(q)) / \tau_g\big)$, with a single hidden-layer MLP. Sparsemax (rather than softmax) yields \emph{exact-zero} entries on irrelevant axes, so unused adapters add no noise. $\tau_g$ anneals from $2.0$ to $0.5$ over training, giving every axis early training signal before the gate sharpens.

\subsection{Training}
\label{sec:method-training}

A single joint stage trains the four LoRA adapters and the gate MLP; the base embedding model remains frozen. The total loss combines a contrastive term on personalized memory and an anchor term on general retrieval:
\begin{equation*}
  \mathcal{L} = \mathcal{L}_{\mathrm{InfoNCE}} + \lambda_{\mathrm{KD}} \cdot \mathcal{L}_{\mathrm{KD}}.
\end{equation*}

\paragraph{Contrastive loss on AxisCF.}
$\mathcal{L}_{\mathrm{InfoNCE}}$ uses $\mathrm{score}(q, m)$ from \S\ref{sec:method-architecture} on AxisCF, our synthetic axis-counterfactual training corpus ($4{,}196$ trips across four memory formats; construction details in Appendix~\ref{sec:appendix-axiscf-generation}). Each training instance (one AxisCF triplet) consists of a positive memory $m^+$, an anchor query $q$ taken verbatim from $m^+$, and $1$--$2$ single-edit axis CFs of $m^+$ (one CF per axis, with $1$--$2$ axes sampled per instance). The batch memory pool $\mathcal{B}$ contains every positive and every CF in the batch, so each $\mathrm{CF}_A$ of $m^+$ acts as a hard negative for $q$ and as an in-batch negative for other queries. The InfoNCE objective at temperature $\tau$ is:
\begin{equation*}
  \mathcal{L}_{\mathrm{InfoNCE}} = -\log \frac{e^{\mathrm{score}(q, m^+) / \tau}}{\sum_{m \in \mathcal{B}} e^{\mathrm{score}(q, m) / \tau}}.
\end{equation*}

\paragraph{Knowledge-distillation anchor on general retrieval.}
The axis term contributes to every query's score at inference, including queries that lie outside the AxisCF training distribution. Without regularization the LoRAs can overfit to AxisCF and produce arbitrary embeddings on general text, injecting noise into the axis term. We anchor each LoRA to the base embedding model on a subset of MS~MARCO \citep{bajaj2018msmarco}:
\begin{equation*}
  \mathcal{L}_{\mathrm{KD}}
  = \frac{1}{|\mathcal{A}|} \sum_{A \in \mathcal{A}}
    \mathbb{E}_{x \sim \mathrm{MS\,MARCO}}
    \big\| \tilde{\boldsymbol{\phi}}_A(x) - \tilde{\mathbf{z}}_b(x) \big\|_2^2.
\end{equation*}
The anchor only acts on MS~MARCO inputs, leaving the LoRAs free to specialize per axis on AxisCF under the contrastive loss.

\section{Experiments}
\label{sec:results}

\subsection{Setup}
\label{sec:results-setup}

\paragraph{Benchmarks.} We target the retrieval step in three personalized-memory QA benchmarks, covering four splits in total: LongMemEval (LME) \citep{wu2024longmemeval}, a long-term memory benchmark with multi-session chat histories ($940$-session memory pool); ATM-Bench \citep{mei2026atmbench}, a multimodal personal-memory benchmark (images, videos, emails) with two splits, Normal and Hard (Hard with more multi-evidence queries), over a $\sim$$11$K-memory pool; and LoCoMo \citep{maharana2024locomo}, very long-term multi-session conversations spanning multiple months ($\sim$$5.9$K-turn memory pool). Per-split statistics and sample queries are in Appendix~\ref{sec:appendix-eval-data}. We report R@10 as the primary metric: the benchmarks pose multi-evidence queries ($1.4$--$6.3$ gold evidences per query on average across the four splits), so smaller cutoffs cannot reflect evidence recall. The full per-metric breakdown (R@1, R@5, R@10, R@100, MRR) is in Appendix~\ref{sec:appendix-per-metric}.

\paragraph{Models.} We consider three standard dense embedding models spanning two architecture families: MiniLM-L6-v2 \citep{wang2020minilm} ($22$\,M, BERT-style), bge-base-en-v1.5 \citep{xiao2024bge} ($110$\,M, BERT-style), and Qwen3-Embedding-0.6B \citep{qwen3embedding2025} ($600$\,M, causal-attention LLM-based). Queries are prefixed following each model's recommended usage. For Qwen3-Embedding-0.6B, which accepts a task instruction, we use a personal-memory retrieval instruction. Full prefixes are in Appendix~\ref{sec:appendix-hp-table}. The main result covers all three models, while remaining experiments default to Qwen3-Embedding-0.6B.

\paragraph{Hyperparameters.} The core Axis-LoRA configuration is shared across the three models: LoRA rank $r = 16$ on the last $20\%$ of attention layers, inference mixing weight $\alpha = 0.25$. Full per-model hyperparameters and their selection process are in Appendix~\ref{sec:appendix-hp-table}.

\subsection{Main result}
\label{sec:results-main}
\begin{table*}[t]
  \centering
  \small
  \setlength{\tabcolsep}{6pt}
  \caption{R@10 across three encoders on LongMemEval (LME), ATM-Bench Normal (ATM-N), ATM-Bench Hard (ATM-H, the multi-evidence split), and LoCoMo. Each encoder block reports \textbf{Base} (the frozen embedding model with no adaptation, deterministic across seeds) and \textbf{Axis-LoRA} (mean${\scriptstyle\pm}$std over $4$ random seeds), with $\Delta$ the per-benchmark gain in pp on the seed mean; the bottom row averages across the four benchmarks. Axis-LoRA delivers $+1.28$ to $+1.56$~pp 4-bench mean R@10 across all three encoders, with the largest per-bench gain $+3.75$~pp on ATM-Hard (Qwen3-0.6B).}
  \label{tab:main-4bench}
  \begin{tabular}{@{}lccc|ccc|ccc@{}}
    \toprule
     & \multicolumn{3}{c|}{\textbf{MiniLM-L6-v2}}
     & \multicolumn{3}{c|}{\textbf{bge-base-en-v1.5}}
     & \multicolumn{3}{c}{\textbf{Qwen3-Embedding-0.6B}} \\
    \cmidrule(lr){2-4} \cmidrule(lr){5-7} \cmidrule(lr){8-10}
    \textbf{Bench} & Base & Axis-LoRA & $\Delta$ & Base & Axis-LoRA & $\Delta$ & Base & Axis-LoRA & $\Delta$ \\
    \midrule
    LME    & $0.4741$ & $\mathbf{0.4872}{\scriptstyle\,\pm0.001}$ & $+1.31$ & $0.6099$ & $\mathbf{0.6397}{\scriptstyle\,\pm0.003}$ & $+2.98$ & $0.7174$ & $\mathbf{0.7238}{\scriptstyle\,\pm0.002}$ & $+0.64$ \\
    ATM-N  & $0.6465$ & $\mathbf{0.6571}{\scriptstyle\,\pm0.002}$ & $+1.06$ & $0.7286$ & $\mathbf{0.7368}{\scriptstyle\,\pm0.000}$ & $+0.82$ & $0.7544$ & $0.7420{\scriptstyle\,\pm0.003}$ & $-1.24$ \\
    ATM-H  & $0.2806$ & $\mathbf{0.3051}{\scriptstyle\,\pm0.003}$ & $+2.45$ & $0.2911$ & $\mathbf{0.3148}{\scriptstyle\,\pm0.008}$ & $+2.37$ & $0.2994$ & $\mathbf{0.3370}{\scriptstyle\,\pm0.017}$ & $\mathbf{+3.75}$ \\
    LoCoMo & $0.4422$ & $\mathbf{0.4564}{\scriptstyle\,\pm0.002}$ & $+1.42$ & $0.6244$ & $0.6178{\scriptstyle\,\pm0.002}$ & $-0.66$ & $0.5856$ & $\mathbf{0.6053}{\scriptstyle\,\pm0.003}$ & $+1.97$ \\
    \midrule
    \textbf{Avg.} & $0.4608$ & $\mathbf{0.4765}{\scriptstyle\,\pm0.001}$ & $\mathbf{+1.56}$ & $0.5635$ & $\mathbf{0.5773}{\scriptstyle\,\pm0.001}$ & $\mathbf{+1.38}$ & $0.5892$ & $\mathbf{0.6020}{\scriptstyle\,\pm0.005}$ & $\mathbf{+1.28}$ \\
    \bottomrule
  \end{tabular}
\end{table*}

Table~\ref{tab:main-4bench} reports per-benchmark R@10. Axis-LoRA consistently improves the 4-benchmark mean on every model: $+1.56$~pp on MiniLM-L6-v2, $+1.38$~pp on bge-base-en-v1.5, and $+1.28$~pp on Qwen3-Embedding-0.6B. The gains are means over benchmarks: two of the twelve (encoder, benchmark) cells regress (Qwen3-0.6B/ATM-Normal $-1.24$~pp, bge-base/LoCoMo $-0.66$~pp); Appendix~\ref{sec:appendix-regression} traces both dips to a narrow rank-$10$ boundary effect. The largest single per-bench lift is on ATM-Bench Hard ($+3.75$~pp on Qwen3-Embedding-0.6B, $+2.45$~pp on MiniLM, $+2.37$~pp on bge-base), positive on all three encoders. These gains are achieved with a lightweight setup: a per-axis adapter ($\leq 0.6\%$ of base parameters) trained on only $4{,}196$ AxisCF triplets with the base embedding model frozen. All numbers are $4$-seed means with cross-seed std $\leq 0.017$ per cell.

The 4-benchmark mean gains are statistically significant for all three encoders (Table~\ref{tab:main-ci}). Query-level bootstrap $95\%$ confidence intervals ($20$k resamples of per-query R@10 deltas, averaged over the $4$ seeds) are $[+0.54, +2.70]$ on MiniLM, $[+0.25, +2.39]$ on bge-base, and $[+0.15, +2.52]$ on Qwen3-0.6B, all excluding zero. The ATM-Hard lift, tested pooled across the three encoders because of the small split size ($n{=}31$ queries), is likewise significant ($+2.86$~pp $[+0.55, +5.36]$). Appendix~\ref{sec:appendix-ci} reports per-split intervals, together with means excluding ATM-Hard to check whether the gains hinge on this single small split.

\begin{table}[t]
  \centering
  \small
  \caption{Query-level bootstrap $95\%$ confidence intervals for the R@10 gains of Table~\ref{tab:main-4bench} ($20$k resamples of per-query deltas, averaged over $4$ seeds). All intervals exclude zero. ATM-Hard ($n{=}31$) is pooled across the three encoders. Per-split intervals are in Appendix~\ref{sec:appendix-ci}.}
  \label{tab:main-ci}
  \begin{tabular}{@{}lcc@{}}
    \toprule
    & $\Delta$ (pp) & $95\%$ CI \\
    \midrule
    \multicolumn{3}{@{}l}{\emph{4-benchmark mean}} \\
    ~~MiniLM-L6-v2     & $+1.56$ & $[+0.54, +2.70]$ \\
    ~~bge-base-en-v1.5 & $+1.38$ & $[+0.25, +2.39]$ \\
    ~~Qwen3-0.6B       & $+1.28$ & $[+0.15, +2.52]$ \\
    \midrule
    ATM-Hard, 3-encoder pooled & $+2.86$ & $[+0.55, +5.36]$ \\
    \bottomrule
  \end{tabular}
\end{table}

\subsection{General-retrieval preservation}
\label{sec:results-mteb}

We verify that Axis-LoRA preserves general-retrieval competence on MTEB \citep{muennighoff2023mteb}, a standard benchmark for text embedding models, evaluated on a 6-task BEIR subset (NFCorpus, SciFact, ArguAna, SCIDOCS, FiQA2018, TRECCOVID) with Qwen3-Embedding-0.6B. The metric is nDCG@10, the standard rank-sensitive retrieval metric used in MTEB.

As shown in Table~\ref{tab:mteb-beir}, mean nDCG@10 with Axis-LoRA is $0.5399$ versus the base's $0.5449$, a $-0.50$~pp shift. The MS~MARCO KD anchor keeps each LoRA branch close to the base on non-axis text without preventing the personalized-memory amplification.

\begin{table}[t]
  \centering
  \footnotesize
  \setlength{\tabcolsep}{4pt}
  \caption{General retrieval on MTEB BEIR-6 (nDCG@10) with Qwen3-Embedding-0.6B. Axis-LoRA preserves general retrieval competence (mean $\Delta = -0.50$~pp).}
  \label{tab:mteb-beir}
  \resizebox{\columnwidth}{!}{%
  \begin{tabular}{@{}lcccccc|c@{}}
    \toprule
    \textbf{Setup} & NFC. & SciF. & ArgA. & SciD. & FiQA & TREC-C & \textbf{Avg.} \\
    \midrule
    Base       & $0.3616$ & $0.6958$ & $0.6790$ & $0.2305$ & $0.4508$ & $0.8516$ & $0.5449$ \\
    Axis-LoRA  & $0.3605$ & $0.6924$ & $0.6744$ & $0.2289$ & $0.4423$ & $0.8410$ & $0.5399$ \\
    \bottomrule
  \end{tabular}}
\end{table}

\subsection{Ablation studies}
\label{sec:results-ablation}

Two complementary ablations isolate the contributions of Axis-LoRA's design. The first (Table~\ref{tab:3way-ablation}) compares Axis-LoRA against four design variants spanning external prompting to internal per-axis LoRA, isolating where and how axis supervision should be applied. The second (Table~\ref{tab:component-ablation}) reports leave-one-out ablations of the three remaining method components. All ablations use Qwen3-Embedding-0.6B with the same 4-bench eval pipeline; tabulated values are $4$-seed means (per-cell stds in Appendix~\ref{sec:appendix-ablation-std}).

\paragraph{Where to inject axis supervision.}
Table~\ref{tab:3way-ablation} progresses from external to internal interventions; we describe each variant alongside its measured effect.

(a) \textbf{Axis-prefix prompt} replaces the default Qwen instruction (\S\ref{sec:results-setup}) with four axis-specific instructions that direct the model to retrieve memories given a question about a particular axis (e.g., the TEMPORAL prefix begins ``\emph{Given a question about WHEN (time, date, period) something happened, retrieve relevant memory entries.}''; full set in Appendix~\ref{sec:appendix-hp-table}). The query is encoded four times, once under each axis prefix, while the memory side is encoded once with no prefix; the score for each memory is the maximum of the four per-axis cosines. We also tried mean aggregation and $5$-cosine variants that include the default-prefix encoding, and all options give similar 4-bench means within $\pm 0.002$. Axis-prefix prompting yields a small lift of $+0.9$~pp over Base, indicating per-axis specification has a real effect. On the multi-evidence ATM-Hard split, however, prompting falls behind Axis-LoRA by $3.1$~pp: \emph{prompt-only treatment cannot reshape the embedding enough}.

(b) \textbf{TF-PCA} identifies axis-specific directions in the embedding space without training. For each axis $A$, we collect how the embedding shifts when axis $A$ is flipped (the residual $\mathbf{e}(M) - \mathbf{e}(\mathrm{CF}_A)$) across many memories, whiten these residuals, and take the top PCA components as the axis-$A$ subspace. (c) \textbf{MLP+InfoNCE} replaces the training-free subspace with a per-axis $2$-layer MLP head $\phi_A$ trained on the same axis CFs via InfoNCE (positives $(q_A, \phi_A(m))$, hard negatives $(q_A, \phi_A(\mathrm{CF}_A))$). At inference, both project query and memory through their four per-axis subspaces/heads and uniformly average the four per-axis cosines (gate routing requires inside-model supervision unavailable to output-side projections). The two output-side projections give only small mean R@10 changes (TF-PCA: $-0.4$~pp, essentially flat; MLP+InfoNCE: $+0.5$~pp): \emph{output-side projections cannot bridge the R@10 gap}, confirming \S\ref{sec:diagnosis-p1}.

(d) \textbf{Shared-trunk LoRA} uses a single shared LoRA inside the embedding model (instead of four per-axis ones as in Axis-LoRA), followed by four axis-specific output projection heads to recover per-axis embeddings. This isolates the contribution of \emph{per-axis} adaptation inside the model. Without per-axis structure inside the model, shared-trunk LoRA collapses to $0.493$ ($-9.6$~pp below Base). Reintroducing per-axis branches with sparsemax gating (Ours) lifts mean R@10 by $+10.9$~pp over shared-trunk: forcing four axis-decomposed outputs through shared weights corrupts the representation, while separate per-axis branches avoid this interference.

\begin{table}[t]
  \centering
  \footnotesize
  \setlength{\tabcolsep}{3pt}
  \caption{Design-variant ablation (R@10 per benchmark). Rows progress from no axis-aware treatment (Base) through axis-aware prompting, training-free and trained external projections, inside-model shared-trunk LoRA, and full Axis-LoRA. MLP+InfoNCE, Shared-trunk LoRA, and Axis-LoRA are $4$-seed means; other rows are deterministic. Per-cell stds in Appendix~\ref{sec:appendix-ablation-std}.}
  \label{tab:3way-ablation}
  \resizebox{\columnwidth}{!}{%
  \begin{tabular}{@{}lcccc|c@{}}
    \toprule
    \textbf{Method} & \textbf{LME} & \textbf{ATM-N} & \textbf{ATM-H} & \textbf{LoCoMo} & \textbf{Avg.} \\
    \midrule
    Base                & $0.717$ & $\mathbf{0.754}$ & $0.299$ & $0.586$ & $0.589$ \\
    \midrule
    Axis-prefix prompt  & $0.723$ & $0.748$ & $0.308$ & $0.603$ & $0.596$ \\
    TF-PCA              & $0.716$ & $0.749$ & $0.285$ & $0.581$ & $0.583$ \\
    MLP+InfoNCE         & $\mathbf{0.726}$ & $0.750$ & $0.300$ & $0.591$ & $0.592$ \\
    Shared-trunk LoRA   & $0.635$ & $0.611$ & $0.255$ & $0.469$ & $0.493$ \\
    \textbf{Axis-LoRA}  & $0.724$ & $0.742$ & $\mathbf{0.337}$ & $\mathbf{0.605}$ & $\mathbf{0.602}$ \\
    \bottomrule
  \end{tabular}}
\end{table}

\paragraph{Component contributions.}
Table~\ref{tab:component-ablation} reports leave-one-out ablations of the three remaining method components: the axis-CF training signal, the KD anchor, and the sparsemax gate. Replacing axis CFs with MS~MARCO negatives drops mean R@10 by $-0.9$~pp, confirming that the axis-CF training signal is essential. Removing the KD anchor drops mean R@10 by $-1.4$~pp, as the LoRAs drift on non-axis text and inject noise into the axis term at inference. Replacing the learned gate with uniform routing at inference drops mean R@10 by $-0.3$~pp, since uniform weighting dilutes the branches the gate has learned to emphasize. Appendix~\ref{sec:appendix-gate-behavior} analyzes the gate's routing behavior in detail.

\begin{table}[t]
  \centering
  \footnotesize
  \setlength{\tabcolsep}{3pt}
  \caption{Component ablations (R@10 per benchmark, $4$-seed means) on the Axis-LoRA training signal, KD anchor, and gate. All three components are essential: removing any drops Avg.\ R@10 below Axis-LoRA. Per-cell stds in Appendix~\ref{sec:appendix-ablation-std}.}
  \label{tab:component-ablation}
  \resizebox{\columnwidth}{!}{%
  \begin{tabular}{@{}lcccc|c@{}}
    \toprule
    \textbf{Setting} & \textbf{LME} & \textbf{ATM-N} & \textbf{ATM-H} & \textbf{LoCoMo} & \textbf{Avg.} \\
    \midrule
    Base                    & $0.717$ &$\mathbf{0.754}$ & $0.299$ & $0.586$ & $0.589$ \\
    \textbf{Axis-LoRA}      & $0.724$ & $0.742$ & $\mathbf{0.337}$ & $\mathbf{0.605}$ & $\mathbf{0.602}$ \\
    \midrule
    $-$ axis-CF supervision & $\mathbf{0.731}$ & $0.751$ & $0.297$ & $0.592$ & $0.593$ \\
    $-$ KD anchor           & $0.720$ & $0.735$ & $0.324$ & $0.575$ & $0.588$ \\
    $-$ gate (uniform routing) & $0.723$ & $0.752$ & $0.316$ & $\mathbf{0.605}$ & $0.599$ \\
    \bottomrule
  \end{tabular}}
\end{table}

\section{Related Work}
\label{sec:related}

\paragraph{Multi-aspect retrieval.} Retrievers map queries and passages to a shared embedding space and are trained with contrastive objectives \citep{karpukhin2020dpr, xiong2021ance, qu2021rocketqa, izacard2022contriever, li2023gte, xiao2024bge}. Multi-aspect retrieval extends this foundation by representing items along multiple informational aspects rather than a single semantic similarity \citep{kong2022multiaspect, sun2024mural}. More recent work moves aspect specification into the query itself, through embedding models that accept a task instruction as part of the input \citep{lee2025nvembed, qwen3embedding2025}. A parallel line targets a single axis directly, predominantly the temporal one \citep{wu2024tsretriever, abdallah2025tempretriever, han-etal-2025-temporal, huynh2026tmrl}. Axis-LoRA extends this line to four axes jointly, placing one adapter per axis inside the embedding model and gating them per query.

\paragraph{Adapter composition and routing.} Low-rank adapters \citep{hu2022lora} have become a standard way to inject specialized behavior into a frozen base model, and composing or routing across multiple LoRA modules has been studied for cross-task generalization \citep{huang2024lorahub, wu2024mole, kim2025domix}. A related approach routes queries among multiple expert retrieval models \citep{lee2025routerretriever}. Axis-LoRA composes one LoRA adapter per axis and routes across them per query.

\paragraph{Personalization.} Personalization of retrieval and language models adapts system behavior to individual users \citep{zhang2025personalization}. Retrieval-side approaches encode user history into embeddings \citep{zeng2023uia, doddapaneni2024uem}, ground retrieval in user-specific memory structures \citep{wang2024memory}, or build conversational user profiles \citep{joko2024sequential}. Parameter-efficient fine-tuning has likewise been adapted to per-user personalization on the language-model side \citep{tan2024profiles, park2026prisp}. We instead target retrieval from a single user's personal memory itself, where adaptation is driven by the memory content rather than by user identity.

\section{Conclusion}
\label{sec:conclusion}

We showed that frozen dense embedding models under-expose the four axes of personal-memory text in an axis-asymmetric pattern, across BERT-style and causal-attention LLM-based architectures. Two probes localize the cause inside the model: external projection cannot bridge the R@10 gap, and each axis draws from a distinct attention-head footprint. Building on this diagnosis, we proposed Axis-LoRA, which combines four per-axis LoRA adapters inside the model, a per-query sparsemax gate, and a general-retrieval KD anchor. Axis-LoRA lifts 4-benchmark mean R@10 by $+1.28$ to $+1.56$~pp across three embedding models, with the largest per-bench gains on ATM-Hard ($+2.4$ to $+3.8$~pp across the three encoders), while training only on a small axis-counterfactual corpus.

\section*{Limitations}

Several aspects of Axis-LoRA's scope and design are left for future work. The four axes (TEMPORAL, SPATIAL, ENTITY, EVENT) are tied to the benchmarks' annotation schemas, and other domains (e.g., clinical, legal) would need their own axis schema and counterfactual construction. AxisCF's counterfactuals edit explicit axis values; implicit cues, such as relative time expressions, vague locations, and entangled multi-entity events, are under-represented. Axis-LoRA itself is a simple per-axis baseline motivated by the diagnostic findings, leaving room for richer routing (mixture-of-experts gates), stronger inside-model supervision, and joint training with retrieval-aware objectives. Finally, we focus on simple, data-efficient adaptation of a frozen embedding model. Whether the same recipe extends to per-user personalization, quickly fitting a retrieval model to a single user from a small amount of their own interactions, is a natural next step.

\section*{Ethical Considerations}

Our work uses personal-memory benchmarks (LongMemEval, ATM-Bench, LoCoMo) released by their original authors under appropriate privacy review; we do not collect any new user data. To check for test-set leakage, our AxisCF training corpus is screened against a contamination blacklist of named entities and date strings mined from the ATM-Bench corpus and a related personal-memory corpus (Appendix~\ref{sec:appendix-axiscf-generation}). AI assistants (ChatGPT, Claude) were used for minor grammar and writing refinement.


\section*{Acknowledgments}
This work was supported in part by the National Research Foundation of Korea (NRF) grant
[No. RS-2025-02263628], the Institute of Information \& communications Technology Planning
\& Evaluation (IITP) grants [RS-2021-II212068, RS-2022-II220113, RS-2022-II220959, RS-2021-II211343], and the BK21 FOUR Education and
Research Program for Future ICT Pioneers (Seoul
National University), funded by the Korean government. It was also supported by Mobile
eXperience (MX) Business, Samsung Electronics Co., Ltd.
\bibliography{custom}

\appendix

\section{AxisCF: synthetic axis-counterfactual corpus}
\label{sec:appendix-axiscf}

AxisCF is the synthetic axis-controlled personal-memory corpus used both for the phenomenon analyses in \S\ref{sec:phenomenon}--\S\ref{sec:diagnosis} and for training Axis-LoRA in \S\ref{sec:method}. Each item is a triplet of a personal-memory text $M$, an anchor query $q$ taken verbatim from $M$, and a set of single-edit axis counterfactuals $\{\mathrm{CF}_A\}$, where each $\mathrm{CF}_A$ alters only the schema-slot values of axis $A$. Since $M$ and its counterfactuals differ only in those slot values, ranking $M$ above them concentrates the contrastive signal on exactly the under-exposed axes.

\subsection{Schema and counterfactual definition}
\label{sec:appendix-axiscf-schema}

\paragraph{Schema and slot edits.} Each memory follows a small slot schema spanning the four axes: TEMPORAL (date, time), SPATIAL (city, venue), ENTITY (person, object), and EVENT (action, event type). A counterfactual $\mathrm{CF}_A$ replaces the slot values of axis $A$ with alternatives (e.g., \emph{Tuesday} $\to$ \emph{Friday}) while preserving every other slot verbatim. Slot values may span multiple tokens, but only one axis changes per CF; we therefore refer to these as \emph{single-edit} counterfactuals in the schema-slot sense.

\paragraph{Per-axis counterfactual generation.} For each positive memory, we generate $K=3$ counterfactuals per axis with a prompt that constrains the change to a single axis-class. The prompt enforces token-class rules: TEMPORAL alters only dates, clock times, and year tokens (e.g., \emph{Aug 15} $\to$ \emph{Dec 22}); SPATIAL alters only place or venue names within a venue-type whitelist (city, hotel, restaurant, cafe, university, $\ldots$); ENTITY alters only a single person, organization, brand, or product within a subcategory whitelist; EVENT alters only an activity term within a tight category whitelist (e.g., \emph{lunch} $\to$ \emph{dinner} within \emph{Meal}, not \emph{lunch} $\to$ \emph{flight}). TEMPORAL further excludes event-referring temporal expressions (e.g., \emph{Christmas}, \emph{birthday}) to prevent latent EVENT leakage through date words. These whitelists ensure that a single axis is varied while the rest of the schema remains semantically on-policy.

\subsection{Generation pipeline}
\label{sec:appendix-axiscf-generation}

\paragraph{Generator and inference settings.} Positive memories and counterfactuals are both generated with Qwen3.5-27B \citep{qwen35} through a vLLM endpoint, run in non-thinking mode (\texttt{enable\_thinking}\,=\,\texttt{False}) at temperature $0.7$. The same model is used at temperature $0.0$ for a separate validation pass (described below).

\paragraph{Memory format diversity.} Memories are sampled across four memory formats that share the axis schema but differ in surface form:
\begin{itemize}
  \setlength{\itemsep}{2pt}
  \item \textbf{Short turn:} a single sentence with one primary axis (TEMPORAL or EVENT).
  \item \textbf{Multi-turn dialogue:} an $8$--$10$-turn chat covering $2$--$3$ axes.
  \item \textbf{Multi-field passage:} an email, image log, or video transcript covering all four axes.
  \item \textbf{Hard composition:} a long-form passage, dialogue, or email thread spanning $3$--$4$ axes.
\end{itemize}
This format mix gives surface variation while keeping the axis-controlled schema constant across forms.

\paragraph{Validation and filtering.} Counterfactual quality is enforced at the prompt level: the generation prompt constrains each CF to a single axis-class edit under the token-class rules above, and $96\%$ of triplets carry at least two counterfactuals per active axis. Triplets with missing counterfactuals are kept and simply train without those hard negatives, against in-batch negatives only. For the evaluation-pool corpus run (Appendix~\ref{sec:appendix-axiscf-eval-pool}), a second pass with Qwen3.5-27B at temperature $0.0$ additionally verifies that each counterfactual changes only the targeted axis and that the substituted value stays within the whitelist. Failing counterfactuals are discarded.

\paragraph{Contamination prevention.} AxisCF generation is independent of the evaluation benchmarks by construction: prompts contain only the axis schema, generic category whitelists, and synthetic personas, and no benchmark text enters the generation pipeline. To audit the residual risk of incidental overlap, we screen the corpus against a contamination blacklist of named entities, location names, and date strings mined from the ATM-Bench corpus and a related personal-memory corpus. Candidate entries are collected by running spaCy \texttt{en\_core\_web\_trf} NER and a date regex over the test-set memories; Qwen3.5-27B then filters the candidates down to genuine named entities, and the list is normalized, deduplicated, and manually reviewed, leaving roughly $76$K entries. The screen flags no benchmark-identifying entity in AxisCF: every overlap with the blacklist's high-precision entries is a generic real-world term (major cities, hotel and electronics brands, holidays, common first names), while entities that identify specific benchmark memories (conference names, unique local venues) do not appear.

\subsection{Evaluation pool}
\label{sec:appendix-axiscf-eval-pool}

\paragraph{Splits.} Two disjoint splits with zero content overlap are used:
\begin{itemize}
  \item \textbf{Training split} ($4{,}196$ trips): used for Axis-LoRA training (\S\ref{sec:method}). A stratified $90$/$10$ train/validation split is taken jointly over memory format and axis combination so that all $14$ (format, axis-set) strata are represented in both folds.
  \item \textbf{Fair evaluation pool} ($662$ axis-anchored positives after a content-presence filter): used as the controlled-pool setting in \S\ref{sec:phenomenon-behavioral} and \S\ref{sec:diagnosis-p1}, and re-used in Appendix~\ref{sec:method-on-sspe2-fair} for an out-of-distribution method evaluation. This pool is generated from a separate corpus run with the same generator and axis schema as the training split, with no shared seeds or content.
\end{itemize}

\paragraph{Axis balancing.} The fair evaluation pool is balanced across the four axes, so per-axis R@10 and per-axis cosine distances reported in \S\ref{sec:phenomenon} reflect equal-weight contributions rather than corpus-frequency artifacts.

\paragraph{Axis-anchored eval queries.} For the axis-anchored retrieval task in \S\ref{sec:phenomenon-behavioral} and \S\ref{sec:diagnosis-p1}, each query is built from one schema value of a positive memory via a per-axis template:
\begin{itemize}
  \item TEMPORAL: \emph{``This event happened on \{value\}.''} (e.g., \emph{``This event happened on 2024-06-15.''})
  \item SPATIAL: \emph{``This took place at \{value\}.''} (e.g., \emph{``This took place at the Boston conference.''})
  \item ENTITY: \emph{``This memory involves \{value\}.''} (e.g., \emph{``This memory involves Marco.''})
  \item EVENT: \emph{``This memory is about \{value\}.''} (e.g., \emph{``This memory is about the rehearsal dinner.''})
\end{itemize}
A content-presence filter keeps only memories whose axis value appears verbatim in the memory text, so the retrieval target is to recover the memory by its actual axis value rather than a paraphrase. Failure on this task therefore does not reflect missing information: the value is in the gold memory's text, but cosine retrieval compares single pooled vectors, and the value's contribution to the memory embedding is only $1$--$14\%$ of the inter-memory distance scale (\S\ref{sec:phenomenon-structural}). This is too small for the ranking to exploit, even though a lexical matcher would find the token directly. Lexical matching does not, however, substitute for the dense pipeline: personal-memory systems rely on embeddings for semantic content matching, and real benchmark queries often express axis cues non-verbatim (e.g., relative dates). Anchor values also recur across the pool (a median of $40$--$108$ memories share a query's value, depending on the axis), so even an exact-match oracle could not push R@10 close to $1$. Assuming random tie-breaking among memories that contain the queried value, the oracle's expected R@10 is roughly $0.26$--$0.43$ depending on the axis. All four axes fall below this bound, but not equally, and TEMPORAL reaches only $9$--$17\%$ of its bound.

\paragraph{Counterfactual filter choices.} Two probes in the paper use the same AxisCF corpus but apply different counterfactual filters, suited to what each probe measures:
\begin{itemize}
  \item \S\ref{sec:phenomenon-structural} (embedding-level cosine distance) and \S\ref{sec:phenomenon-behavioral} (axis-anchored R@10) use the full per-axis counterfactual set, which allows axis-phrase swaps that may span multiple tokens (e.g., \emph{Boston conference} $\to$ \emph{Seattle library} for SPATIAL). The cosine and retrieval signals are sentence-level (pooled embedding), and an axis-phrase swap is a natural unit of axis change.
  \item \S\ref{sec:diagnosis-p2} (head-spectrum probe) restricts to counterfactuals that alter exactly one schema-slot token, because per-(layer, head) responses are per-token signals that mix when multiple tokens change at once. The strict filter shrinks the SPATIAL pool to $n=41$; a relaxed cross-check ($n=254$) and a position-controlled probe confirm the qualitative ordering, see Appendix~\ref{sec:appendix-head-spectrum-robustness}.
\end{itemize}
Under the strict single-token filter, the \S\ref{sec:phenomenon-structural} cosine ratios shift somewhat (SPATIAL becomes a smaller multiple of the edit floor, since a single-token swap is a smaller perturbation than a full phrase swap), but the qualitative ordering (TEMPORAL the smallest, the other three above the floor) is unchanged.

\section{Phenomenon supporting analyses}
\label{sec:appendix-phenomenon-supporting}

\subsection{Axis-blindness reference scales}
\label{sec:appendix-axis-blindness}

This appendix provides the reference scales used by Figure~\ref{fig:axis-blindness} and Table~\ref{tab:axis-scale-graded} in \S\ref{sec:phenomenon-structural}.

\paragraph{Non-axis 1-word edit floor.} For each encoder, the edit floor is the mean cosine distance between a memory $M$ and a single-token control edit on a non-axis content word (axis values excluded). It reports the response magnitude to a near-noise content perturbation and serves as a per-encoder lower bound for ``meaningful'' axis response. Per-encoder values: MiniLM $f_e\!=\!0.023$, bge-base $f_e\!=\!0.012$, Qwen3-Embedding-0.6B $f_e\!=\!0.010$ (Table~\ref{tab:axis-scale-graded}, top block).

\paragraph{Inter-memory distance $\mu_{\mathrm{inter}}$.} For each model, $\mu_{\mathrm{inter}}$ is the mean cosine distance between any two memories in the corpus. It reflects the operating scale at which cosine retrieval distinguishes candidates: anything far below this scale will not be discriminated by cosine ranking. Per-model values: MiniLM $\mu_{\mathrm{inter}}\!=\!0.78$, bge-base $\mu_{\mathrm{inter}}\!=\!0.47$, Qwen3-Embedding-0.6B $\mu_{\mathrm{inter}}\!=\!0.65$. Across all three models the per-axis CF cosine distances sit at $1$--$13\%$ of $\mu_{\mathrm{inter}}$, confirming that axis swaps fall far below the inter-memory scale that determines retrieval rank.

\subsection{Robustness checks}
\label{sec:appendix-phenomenon-controls}

We address three alternative explanations for the axis-asymmetric under-exposure documented in \S\ref{sec:phenomenon}, and one caveat on why the phenomenon analyses use a controlled synthetic pool rather than real benchmarks.

\paragraph{Edit-magnitude control.} A natural concern is whether TEMPORAL appears smaller simply because TEMPORAL edits modify fewer tokens than other axes. Single-edit-per-axis counterfactuals already match the schema-slot edit count, but we further normalize cosine response by Jaccard token-edit distance. On Qwen3-Embedding-0.6B (single-edit aggregate):

\begin{table}[h]
  \centering
  \footnotesize
  \caption{Edit-magnitude control on Qwen3-Embedding-0.6B (single-edit aggregate): per-axis cosine-response ratios vs.\ TEMPORAL, raw and after Jaccard token-edit normalization.}
  \label{tab:appendix-jaccard-ratios}
\begin{tabular}{lcc}
\toprule
Ratio & Raw & Jaccard-normalized \\
\midrule
ENTITY / TEMPORAL  & $4.92$ & $1.76$ \\
SPATIAL / TEMPORAL & $7.56$ & $2.09$ \\
EVENT / TEMPORAL   & $4.82$ & $1.92$ \\
\bottomrule
\end{tabular}
\end{table}

TEMPORAL remains uniquely small after the strictest edit-magnitude control; roughly $25$--$30\%$ of the raw asymmetry is edit-magnitude artifact, but $70$--$75\%$ is encoder-intrinsic.

\paragraph{Position-bias control.} Another concern is that TEMPORAL tokens tend to appear in late prepositional-phrase positions, and encoders may attend less strongly to such positions. We control this by generating five positional variants per memory ($\text{natural}$, TEMPORAL-as-subject, TEMPORAL-as-object, ENTITY-as-subject, ENTITY-as-object) using Qwen3.5-27B, preserving the schema values, and rerun cosine-distance measurements. ENTITY-to-TEMPORAL swap-distance ratio per (model, variant):

\begin{table}[h]
  \centering
  \footnotesize
  \caption{Position-bias control: range of ENTITY-to-TEMPORAL swap-distance ratio across five positional variants (natural, TEMPORAL-as-subject, TEMPORAL-as-object, ENTITY-as-subject, ENTITY-as-object).}
  \label{tab:appendix-position-bias}
\begin{tabular}{lc}
\toprule
Backbone & E / T ratio range across variants \\
\midrule
MiniLM      & $20.4\times$--$40.4\times$ \\
Qwen3-0.6B  & $18.2\times$--$46.2\times$ \\
Qwen3-8B    & $12.6\times$--$31.8\times$ \\
\bottomrule
\end{tabular}
\end{table}

Even when TEMPORAL is moved into the subject position, ENTITY-swap distance remains $\geq 12\times$ larger than TEMPORAL-swap distance. Position is at most a secondary modulator; the asymmetry is not a positional artifact.

\paragraph{Small-edit triviality.} A third concern is that the cosine response to any small token edit is near-noise, in which case TEMPORAL-at-floor is unremarkable. We already report the non-axis 1-word edit floor in \S\ref{sec:phenomenon-structural} and Appendix~\ref{sec:appendix-axis-blindness}: TEMPORAL sits at $0.3$--$1.1\times$ this floor on all three models, while SPATIAL, ENTITY, and EVENT sit at $2.6$--$8.0\times$ — orders of magnitude apart, ruling out the small-edit-triviality reading.

\paragraph{Caveat on real-corpus axis-anchored R@10.} The phenomenon analyses use a synthetic controlled pool rather than real benchmarks (LongMemEval, ATM-Bench, LoCoMo). The reason is that LLM-extracted TEMPORAL values in real conversational corpora are systematically vaguer than other axis values. On LongMemEval, $\sim 32\%$ of extracted TEMPORAL anchors are vague relative phrases (\emph{``next weekend''}, \emph{``tonight''}), $\sim 25\%$ are month-only without year, and only $\sim 5\%$ are full year-dates; in contrast, LongMemEval SPATIAL, ENTITY, and EVENT anchors are mostly specific proper nouns or distinctive activity terms. A real-corpus axis-anchored test would therefore confound encoder under-encoding (our target) with inherent natural-language vagueness of TEMPORAL. We use the controlled pool for the phenomenon claim, and validate end-to-end retrieval on the real benchmarks in \S\ref{sec:results}.

\section{Diagnostic probes: details and robustness}
\label{sec:appendix-diagnostic-probes}

\subsection{Projection probe: training and full results}
\label{sec:appendix-projection-probe}

The projection probe in \S\ref{sec:diagnosis-p1} tests whether axis information that is under-exposed in the raw cosine geometry can be recovered by an output-side head. We considered two variants of increasing capacity; the per-axis MLP with InfoNCE is the variant reported in \S\ref{sec:diagnosis-p1}.

\paragraph{Linear ridge regression.} Per-axis linear map $W_A : \mathbb{R}^d \to \mathbb{R}^d$ fit by ridge regression on $(m, q_A)$ pairs ($\lambda = 1.0$, $5$-fold trip-level CV), evaluated as $\cos(q_A, W_A m)$. This serves as a simple, closed-form baseline. The raw R@10 column matches Table~\ref{tab:axis-anchored-fair}; the projected column reports ridge's lift on the same axis-anchored evaluation. Table~\ref{tab:appendix-projection-ridge} shows that linear ridge lifts TEMPORAL only modestly ($+0.6$--$+3.9$~pp) and is destructive on SPATIAL and ENTITY across all three models, indicating that a single linear map cannot bridge the raw-cosine gap.

\begin{table}[t]
  \centering
  \footnotesize
  \setlength{\tabcolsep}{3pt}
  \caption{Linear ridge variant: per-(model, axis) raw $\to$ projected R@10, $\Delta = $ projected $-$ raw.}
  \label{tab:appendix-projection-ridge}
  \resizebox{\columnwidth}{!}{%
  \begin{tabular}{@{}lccc|ccc|ccc@{}}
    \toprule
     & \multicolumn{3}{c|}{\textbf{MiniLM}} & \multicolumn{3}{c|}{\textbf{bge-base}} & \multicolumn{3}{c}{\textbf{Qwen3-0.6B}} \\
    \textbf{Axis} & raw & proj & $\Delta$ & raw & proj & $\Delta$ & raw & proj & $\Delta$ \\
    \midrule
    TEMP.  & $0.061$ & $0.082$ & $+2.1$ & $0.072$ & $0.078$ & $+0.6$ & $0.039$ & $0.078$ & $+3.9$ \\
    SPAT.  & $0.248$ & $0.182$ & $\mathbf{-6.6}$ & $0.305$ & $0.226$ & $\mathbf{-7.9}$ & $0.231$ & $0.193$ & $\mathbf{-3.8}$ \\
    ENT.   & $0.163$ & $0.101$ & $\mathbf{-6.2}$ & $0.154$ & $0.111$ & $\mathbf{-4.3}$ & $0.152$ & $0.113$ & $\mathbf{-3.9}$ \\
    EVENT  & $0.254$ & $0.261$ & $+0.7$ & $0.295$ & $0.328$ & $+3.3$ & $0.312$ & $0.315$ & $+0.3$ \\
    \bottomrule
  \end{tabular}}
\end{table}

\paragraph{Per-axis MLP with InfoNCE (main).} A per-axis 2-layer MLP $\phi_A$ with shape $d \to 2d \to d$ and GELU. The InfoNCE objective uses three negative types per axis: the positive $(q_A, \phi_A(m))$, the same-trip single-edit axis-$A$ counterfactual as a hard negative $(q_A, \phi_A(\mathrm{CF}_A))$, and in-batch other-trip memories as easy negatives. We optimize each head with Adam (learning rate $1 \times 10^{-4}$), batch size $64$, $30$ epochs, and softmax temperature $\tau = 0.05$. Hyperparameters were not separately tuned per model. We use $5$-fold trip-level cross-validation so each query is scored by a head trained without that trip; the evaluation protocol matches \S\ref{sec:phenomenon-behavioral} (axis-anchored R@10 over the evaluation set). The full $12$-cell results from this variant are reported in Table~\ref{tab:axis-anchored-fair} of the body.

The MLP variant lifts TEMPORAL by $+8.9$--$+14.4$~pp (vs.\ $+0.6$--$+3.9$~pp for linear ridge), while non-TEMPORAL cells either stay near raw or go negative on $3/9$ cells. The MLP is therefore the strongest output-side probe of the two, and is the variant reported in the body.

\subsection{Head-spectrum probe: filtering and robustness}
\label{sec:appendix-head-spectrum-robustness}

The head-spectrum probe in \S\ref{sec:diagnosis-p2} uses a stricter axis-CF filter than the axis-anchored retrieval evaluation in \S\ref{sec:phenomenon-behavioral}: each $(M, \mathrm{CF}_A)$ pair must alter exactly one schema-slot token of axis $A$, not just one slot. Per-axis pair counts under this strict filter: TEMPORAL $n=2191$, EVENT $n=934$, ENTITY $n=266$, SPATIAL $n=41$. SPATIAL is small because raw SPATIAL counterfactuals typically swap both venue and city tokens jointly, and only $41$ pairs survive the strict single-token filter.

\paragraph{SPATIAL $n=41$ cross-checks.} The absolute SPATIAL coverage at $n=41$ has wider sampling uncertainty than the other three axes. Two checks support the qualitative ordering (TEMPORAL most diffuse, the other three more concentrated):
\begin{itemize}
  \item \textbf{Rank ordering.} Replacing absolute \%-of-heads with its rank within the four axes per model, TEMPORAL is the most-diffuse rank-1 in all three models, while SPATIAL, ENTITY, and EVENT permute among ranks 2--4 without any of them dominating.
  \item \textbf{Relaxed-CF reproducibility.} Replacing the strict single-edit SPATIAL CF pool with a relaxed pool that allows venue $+$ city joint swap ($n=254$) slightly inflates the SPATIAL response but preserves the ordering: SPATIAL stays in the concentrated cluster (within $2$~pp of ENTITY/EVENT \%-of-heads) and TEMPORAL stays the most diffuse.
\end{itemize}
Bootstrap $95\%$ CIs on Qwen3-Embedding-0.6B ($1000$ resamples): TEMPORAL $32.1 \pm 1.0\%$, EVENT $25.4 \pm 1.5\%$, ENTITY $18.5 \pm 2.1\%$, SPATIAL $19.6 \pm 4.2\%$ (wider, as expected from the small $n$). TEMPORAL is non-overlapping with the other axes at $95\%$.

\paragraph{Position-controlled probing.} A natural concern is that TEMPORAL tokens may sit at canonical sequence-header positions in personal-memory data, so the diffuse-head pattern could reflect position-based attention routing. We restrict the CF pool to pairs where the perturbed token is at mid-sequence positions ($20$--$100$, past typical header content) and re-measure heads-to-$50\%$-mass.

\begin{table}[h]
  \centering
  \footnotesize
  \setlength{\tabcolsep}{4pt}
  \caption{Position-controlled head-spectrum probe: ENTITY (E) vs.\ TEMPORAL (T) heads-to-$50\%$-mass on the full CF pool vs.\ the mid-sequence subset (perturbed token at positions $20$--$100$).}
  \label{tab:appendix-head-position-controlled}
\resizebox{\columnwidth}{!}{%
\begin{tabular}{lcc}
\toprule
Model & Full pool E\,/\,T \% & Mid-position pool E\,/\,T \% \\
\midrule
MiniLM        & $26.4 / 41.7$ (ratio $0.63$) & $28.1 / 40.9$ (ratio $0.69$) \\
Qwen3-0.6B    & $18.5 / 32.1$ (ratio $0.58$) & $19.4 / 32.4$ (ratio $0.60$) \\
\bottomrule
\end{tabular}}
\end{table}

The ENTITY-to-TEMPORAL ratio stays within $0.07$ of the full-pool value at both models, and on Qwen3-Embedding-0.6B the TEMPORAL-vs-ENTITY gap is $\geq 13$~pp at both pool definitions. The axis-asymmetric head distribution is not a positional artifact.

\section{Method ablations}
\label{sec:appendix-method-ablations}

\subsection{$\alpha$-sensitivity sweep}
\label{sec:alpha-sweep}

The $\alpha$ knob (axis-term mixing weight) sweeps the composite from frozen-base ($\alpha = 0$) to axis-only. 4-bench mean R@10 on Qwen3-Embedding-0.6B stays at or above the frozen base across $\alpha \in [0.1, 0.5]$, with the best value $\mathbf{0.6020}$ at $\alpha=0.25$ versus $0.5892$ at $\alpha=0$. The val winner (val\_axis R@10 $= 0.9658$ at $\alpha=0.25$) coincides with the 4-bench test winner, so $\alpha$ is not test-tuned. Above $\alpha=0.25$, content-overlap-dominant queries (ATM-Normal, LoCoMo) get over-weighted on the axis term and the mean drops, with the axis-only setting ($\alpha=1.0$) collapsing on ATM-Normal.

\subsection{Gate variant: sparsemax vs.\ softmax with axis supervision}
\label{sec:appendix-gate-ce}

Axis-LoRA's per-query gate uses sparsemax and is trained only by the composite contrastive objective. A natural alternative is to replace sparsemax with softmax and add an explicit axis-supervision term — for each training sample with axis label set $A^* \subseteq \{\textsc{T}, \textsc{S}, \textsc{En}, \textsc{Ev}\}$, a cross-entropy loss between the gate weights and a uniform distribution over $A^*$:
\begin{equation*}
  \mathcal{L}_{\mathrm{gate\text{-}CE}} \;=\; - \tfrac{1}{|A^*|}\!\!\sum_{A \in A^*} \log w_A(q),
\end{equation*}
combined as $\mathcal{L} + \lambda_{\mathrm{gate}}\,\mathcal{L}_{\mathrm{gate\text{-}CE}}$ with $\lambda_{\mathrm{gate}} = 0.05$. Every other hyperparameter matches the paper main configuration (\S\ref{sec:results-setup}). Table~\ref{tab:gate-ce-vs-sparsemax} compares the two on bge-base-en-v1.5 over the same $4$ random seeds.

\begin{table}[t]
  \centering
  \footnotesize
  \setlength{\tabcolsep}{4pt}
  \caption{$4$-seed mean${\scriptstyle\pm}$std R@10 on bge-base-en-v1.5. The two gate variants give comparable 4-bench means (difference $0.24$~pp, within per-bench seed variance) with different per-bench trade-offs: sparsemax is slightly stronger on hard-distractor splits (LME, ATM-H), softmax+axis-CE is slightly stronger on the longer-context splits (ATM-N, LoCoMo).}
  \label{tab:gate-ce-vs-sparsemax}
  \resizebox{\columnwidth}{!}{%
  \begin{tabular}{@{}lcccc|c@{}}
    \toprule
    \textbf{Gate} & \textbf{LME} & \textbf{ATM-N} & \textbf{ATM-H} & \textbf{LoCoMo} & \textbf{Avg.} \\
    \midrule
    Sparsemax (Ours)   & $\mathbf{0.6397}{\scriptstyle\,\pm0.003}$ & $0.7368{\scriptstyle\,\pm0.000}$ & $\mathbf{0.3148}{\scriptstyle\,\pm0.008}$ & $0.6178{\scriptstyle\,\pm0.002}$ & $\mathbf{0.5773}$ \\
    Softmax + axis-CE  & $0.6360{\scriptstyle\,\pm0.001}$ & $\mathbf{0.7405}{\scriptstyle\,\pm0.002}$ & $0.2975{\scriptstyle\,\pm0.003}$ & $\mathbf{0.6256}{\scriptstyle\,\pm0.002}$ & $0.5749$ \\
    \bottomrule
  \end{tabular}}
\end{table}

We keep sparsemax as the paper-main configuration because it is the simpler training recipe: it does not require per-sample axis labels and trains from axis-counterfactual pairs alone, while still producing exact zeros on inactive axes (so off-axis branches add no noise at inference).

\subsection{Gate behavior: routing collapse, a single-adapter control, and axis-supervised routing}
\label{sec:appendix-gate-behavior}

This appendix analyzes what the learned gate of \S\ref{sec:method-architecture} actually does during and after training, reports a single-adapter control that isolates the source of the mean gain, and shows that a light axis supervision on the gate realizes the per-query routing that motivates the design.

\paragraph{The free gate collapses to one branch per encoder.}
Although the gate is queried per example, its learned weights degenerate: averaged over queries, a single axis branch receives nearly all routing mass, and which branch wins varies across random seeds (on Qwen3-Embedding-0.6B, the four paper seeds collapse onto EVENT, SPATIAL, ENTITY, and ENTITY respectively). The collapsed gate carries no per-query signal, giving time-labeled questions no more TEMPORAL weight than other questions (Table~\ref{tab:gate-align}). The gate is still not vestigial at inference: on Qwen3-0.6B, routing with the collapsed learned gate beats uniform mixing over the same branches ($+1.28$ vs.\ $+0.98$~pp 4-benchmark mean over Base), because it selects exactly the branch that absorbed the training signal.

\paragraph{Why training admits correct routing but does not select it.}
The collapse is not due to a missing learning signal. Writing the training score as $\mathrm{score}(q,m) = \mathrm{sim}_b(q,m) + \alpha \sum_A w_A(q)\,\mathrm{sim}_A(q,m)$, the InfoNCE gradient with respect to $w_A(q)$ is proportional to the margin $\mathrm{sim}_A(q, m^+) - \mathrm{sim}_A(q, \mathrm{CF})$, i.e., to how well branch $A$ separates the positive from the axis-labeled counterfactual of that example. An axis-aligned assignment, where each branch specializes on its axis and the gate routes accordingly, is therefore an optimum of the objective. It is, however, not the optimum that training reaches. All branches start identical (LoRA initialization), so the gate gradient initially distinguishes nothing; as soon as one branch pulls ahead, its margins grow on counterfactuals of \emph{every} axis, the gate shifts mass to it on all examples, and the branch monopolizes the gradient. This winner-take-all dynamic mirrors the routing collapse documented for mixture-of-experts training, and the gate temperature annealing of \S\ref{sec:method-architecture} (designed to give every branch early signal) delays but does not prevent it: keeping the final temperature high ($\tau_g \to 1.0$ instead of $0.5$) still collapses onto a single axis on every seed ($+1.09{\scriptstyle\,\pm0.34}$~pp).

\paragraph{A single-adapter control isolates the source of the mean gain.}
The collapsed configuration effectively scores with one jointly-trained branch. We make this exact with a control that routes every training example to a single branch (the other branches remain at the KD anchor) and scores with that branch alone, i.e., one adapter, no per-axis structure, no gate, trained on the full AxisCF set. Table~\ref{tab:gate-decomp} compares 4-benchmark mean gains over Base across the three encoders ($4$ seeds).

\begin{table}[t]
  \centering
  \footnotesize
  \setlength{\tabcolsep}{4pt}
  \caption{Gate analysis, $\Delta$ 4-benchmark mean R@10 (pp) over Base, $4$-seed means. \textbf{Top:} single-adapter control vs.\ Axis-LoRA vs.\ the axis-supervised gate (the loss of Table~\ref{tab:gate-ce-vs-sparsemax} at $\lambda_{\mathrm{gate}}{=}0.05$ with sparsemax retained), with the supervised branches also read out under uniform routing. \textbf{Bottom:} routing decomposition on Qwen3-0.6B.}
  \label{tab:gate-decomp}
  \resizebox{\columnwidth}{!}{%
  \begin{tabular}{@{}lccc@{}}
    \toprule
    & MiniLM & bge-base & Qwen3-0.6B \\
    \midrule
    Single adapter (control) & $+1.50$ & $+1.16$ & $+1.59$ \\
    Axis-LoRA (paper)        & $+1.56$ & $\mathbf{+1.38}$ & $+1.28$ \\
    ~+ axis-sup.\ gate       & $+1.49$ & $+0.93$ & $+1.54$ \\
    ~~~\emph{same, uniform routing} & $+1.55$ & $+1.09$ & $\mathbf{+1.63}$ \\
    \midrule
    & \multicolumn{3}{c}{Qwen3-0.6B routing decomposition} \\
    \midrule
    free gate: learned / uniform    & \multicolumn{3}{c}{$+1.28$ ~/~ $+0.98$} \\
    axis-sup.: learned / uniform    & \multicolumn{3}{c}{$+1.54$ ~/~ $+1.63$} \\
    hard label routing / single adapter & \multicolumn{3}{c}{$+1.35$ ~/~ $+1.59$} \\
    \bottomrule
  \end{tabular}}
\end{table}

The single adapter matches the cross-encoder mean of the full method ($+1.42$ vs.\ $+1.41$~pp). The mean improvement is therefore driven by the axis-counterfactual training signal and adapter capacity rather than by routing per se (replacing the axis counterfactuals with generic negatives erases the full method's gain, Table~\ref{tab:component-ablation}), consistent with the phrasing of \S\ref{sec:results-ablation}. The gated per-axis structure matches the control at no average cost while adding what a single adapter cannot provide: encoder-level axis selection where it pays off (on bge-base the full method leads the control by $+0.22$~pp overall and by $+0.93$~pp on LongMemEval), per-axis representations (the axis-anchored evaluations of Appendix~\ref{sec:method-on-sspe2-fair} require per-axis branches), and a routing space that supervision can activate, as follows.

\paragraph{Axis supervision realizes per-query routing.}
AxisCF already labels which axis each training example edits. Adding the gate cross-entropy of Table~\ref{tab:gate-ce-vs-sparsemax} toward these labels with a small weight ($\lambda_{\mathrm{gate}} = 0.05$), while keeping the paper's sparsemax gate, changes the training dynamics qualitatively: the routing mass spreads over all four branches with a seed-stable distribution (TEMPORAL $\approx .43$, EVENT $\approx .28$, ENTITY $\approx .16$, SPATIAL $\approx .13$ on Qwen3-0.6B), and genuine per-query behavior appears. Time-labeled questions receive measurably more TEMPORAL weight than other questions on LoCoMo for all three encoders and on LME for Qwen3-0.6B (Table~\ref{tab:gate-align}); MiniLM shows no alignment on LME, the same cell where the supervised gate trails uniform routing. On Qwen3-0.6B the supervised gate lifts the mean to $+1.54$ ($+1.63$ with uniform readout) while halving the single regressing cell (ATM-Normal $-1.24 \to -0.67$~pp). Teacher-forcing the labels as hard one-hot routing, which trains each branch in isolation, recovers less of the gain ($+1.35{\scriptstyle\,\pm0.25}$ vs.\ $+1.63{\scriptstyle\,\pm0.16}$ for the soft supervised mixture), suggesting branches benefit from learning to cooperate under the mixed routing used at inference. The effect of supervision is encoder-dependent, mirroring the encoder-level specialization above: it helps Qwen3-0.6B, ties MiniLM, and trades away bge-base's specialization advantage ($+0.93$ vs.\ $+1.38$~pp). We therefore report it as an analysis variant rather than a change to the method.

\begin{table}[t]
  \centering
  \footnotesize
  \setlength{\tabcolsep}{5pt}
  \caption{Per-query alignment: mean TEMPORAL gate weight of the axis-supervised gate on time-labeled questions vs.\ all other questions ($4$-seed means). The freely trained gate shows no time-vs-other differential on any encoder ($\leq .006$); the supervised gate does on LoCoMo for all three encoders and on LME for Qwen3-0.6B.}
  \label{tab:gate-align}
  \resizebox{\columnwidth}{!}{%
  \begin{tabular}{@{}lcccc@{}}
    \toprule
    & \multicolumn{2}{c}{LoCoMo (\emph{temporal}, $n{=}320$)} & \multicolumn{2}{c}{LME (\emph{temp.-reasoning}, $n{=}133$)} \\
    \cmidrule(lr){2-3} \cmidrule(lr){4-5}
    & time & other & time & other \\
    \midrule
    MiniLM     & $\mathbf{.308}$ & $.186$ & $.601$ & $.603$ \\
    bge-base   & $\mathbf{.483}$ & $.256$ & $.653$ & $.627$ \\
    Qwen3-0.6B & $\mathbf{.408}$ & $.338$ & $\mathbf{.682}$ & $.637$ \\
    \bottomrule
  \end{tabular}}
\end{table}

\paragraph{Summary.}
The free gate degenerates into an encoder-level axis selector; this costs nothing on average, and its selection is what the learned-vs-uniform ablation of Table~\ref{tab:component-ablation} measures. The mean gain itself comes from training on joint four-axis counterfactuals, which a single adapter can absorb. The per-axis gated structure earns its place by matching that control on average while enabling encoder-level specialization, per-axis representations, and, under light axis supervision, the per-query routing that motivates the design.

\subsection{Axis-LoRA on the AxisCF fair eval pool}
\label{sec:method-on-sspe2-fair}

The phenomenon (\S\ref{sec:phenomenon-behavioral}) and diagnosis (\S\ref{sec:diagnosis-p1}) sections both evaluate on the AxisCF fair pool ($662$ positives, single axis-anchor + content-presence filter). For consistency we report Axis-LoRA's R@10 on the same pool, scored with the composite $\langle z_b(q), z_b(m)\rangle + \alpha \sum_A w_A(q) \langle \phi_A(q), \phi_A(m)\rangle$ at $\alpha=0.25$. The AxisCF fair pool is generated from a separate corpus run from the training data with $0$ content overlap, so this is an out-of-distribution evaluation of the trained adapter.

\begin{table}[t]
  \centering
  \footnotesize
  \setlength{\tabcolsep}{2.5pt}
  \caption{Axis-anchored R@10 and median rank on the AxisCF fair pool: raw cosine, MLP+InfoNCE projection (\S\ref{sec:diagnosis-p1}), and Axis-LoRA ($\alpha=0.25$, seed 42). Lower median rank = better. Axis-LoRA improves R@10 over raw on $10/12$ (encoder, axis) cells and improves median rank on all $12$ cells, including the three cells where MLP+InfoNCE was destructive (bge-base SPATIAL/EVENT, Qwen3-Embedding-0.6B EVENT).}
  \label{tab:method-sspe2-fair}
  \resizebox{\columnwidth}{!}{%
  \begin{tabular}{@{}llcccccc@{}}
    \toprule
     & & \multicolumn{2}{c}{\textbf{MiniLM}} & \multicolumn{2}{c}{\textbf{bge-base}} & \multicolumn{2}{c}{\textbf{Qwen3-0.6B}} \\
    \cmidrule(lr){3-4} \cmidrule(lr){5-6} \cmidrule(lr){7-8}
    \textbf{Axis} & \textbf{Setup} & R@10 & med & R@10 & med & R@10 & med \\
    \midrule
    \multirow{3}{*}{TEMP}  & raw    & $0.061$ & $157$ & $0.072$ & $109$ & $0.039$ & $168$ \\
                          & MLP    & $0.150$ & $64$  & $0.167$ & $60$  & $0.183$ & $62$  \\
                          & A-LoRA & $0.066$ & $109$ & $\mathbf{0.181}$ & $\mathbf{45}$ & $0.115$ & $82$ \\
    \midrule
    \multirow{3}{*}{SPAT}  & raw    & $0.248$ & $55$  & $0.305$ & $29$  & $0.231$ & $58$  \\
                          & MLP    & $0.250$ & $57$  & $0.275$ & $42$  & $0.233$ & $48$  \\
                          & A-LoRA & $\mathbf{0.304}$ & $\mathbf{34}$ & $\mathbf{0.335}$ & $\mathbf{26}$ & $\mathbf{0.282}$ & $\mathbf{33}$ \\
    \midrule
    \multirow{3}{*}{ENT}   & raw    & $0.163$ & $63$  & $0.154$ & $81$  & $0.152$ & $65$  \\
                          & MLP    & $0.224$ & $65$  & $0.179$ & $73$  & $0.197$ & $49$  \\
                          & A-LoRA & $0.190$ & $58$  & $0.167$ & $82$  & $0.196$ & $50$  \\
    \midrule
    \multirow{3}{*}{EVENT} & raw    & $0.254$ & $52$  & $0.295$ & $32$  & $0.312$ & $39$  \\
                          & MLP    & $0.260$ & $27$  & $0.266$ & $31$  & $0.231$ & $29$  \\
                          & A-LoRA & $0.220$ & $40$  & $\mathbf{0.295}$ & $\mathbf{24}$ & $\mathbf{0.340}$ & $\mathbf{23}$ \\
    \bottomrule
  \end{tabular}}
\end{table}

Table~\ref{tab:method-sspe2-fair} reports the per-axis breakdown. Axis-LoRA improves R@10 over raw on $10/12$ cells (the exceptions are MiniLM EVENT and bge-base EVENT, the latter essentially tied with raw) and improves median rank on all $12$ cells. On the three cells where MLP+InfoNCE was destructive (bge-base SPATIAL/EVENT, Qwen3-Embedding-0.6B EVENT), Axis-LoRA instead lifts or matches raw ($+3.0$/$0.0$/$+2.8$~pp R@10), confirming that inside-encoder reshaping reaches geometry the output-side projection cannot. MiniLM lifts are smaller (the MLP+InfoNCE probe outperforms Axis-LoRA on its TEMPORAL and ENTITY cells), consistent with the small-encoder pattern in \S\ref{sec:results-main} where MiniLM's lift on real benchmarks is also the smallest of the three encoders.

\subsection{Per-cell stds for ablation tables}
\label{sec:appendix-ablation-std}

Table~\ref{tab:ablation-std} gives the $4$-seed mean${\scriptstyle\pm}$std for each cell in Tables~\ref{tab:3way-ablation} and~\ref{tab:component-ablation} that was trained (deterministic rows and the inference-time uniform-routing row are omitted).

\begin{table}[h]
  \centering
  \footnotesize
  \setlength{\tabcolsep}{3pt}
  \caption{$4$-seed mean${\scriptstyle\pm}$std for the trained rows of Tables~\ref{tab:3way-ablation} and~\ref{tab:component-ablation}.}
  \label{tab:ablation-std}
  \resizebox{\columnwidth}{!}{%
  \begin{tabular}{@{}lccccc@{}}
    \toprule
    \textbf{Variant} & \textbf{LME} & \textbf{ATM-N} & \textbf{ATM-H} & \textbf{LoCoMo} & \textbf{Avg.} \\
    \midrule
    Axis-LoRA         & $.724{\scriptstyle\,\pm.002}$ & $.742{\scriptstyle\,\pm.003}$ & $.337{\scriptstyle\,\pm.017}$ & $.605{\scriptstyle\,\pm.003}$ & $.602{\scriptstyle\,\pm.005}$ \\
    Shared-trunk      & $.635{\scriptstyle\,\pm.021}$ & $.611{\scriptstyle\,\pm.011}$ & $.255{\scriptstyle\,\pm.007}$ & $.469{\scriptstyle\,\pm.014}$ & $.493{\scriptstyle\,\pm.008}$ \\
    MLP+InfoNCE       & $.726{\scriptstyle\,\pm.002}$ & $.750{\scriptstyle\,\pm.002}$ & $.300{\scriptstyle\,\pm.004}$ & $.591{\scriptstyle\,\pm.001}$ & $.592{\scriptstyle\,\pm.002}$ \\
    \midrule
    $-$ axis-CF supv. & $.731{\scriptstyle\,\pm.004}$ & $.751{\scriptstyle\,\pm.003}$ & $.297{\scriptstyle\,\pm.003}$ & $.592{\scriptstyle\,\pm.003}$ & $.593{\scriptstyle\,\pm.002}$ \\
    $-$ KD anchor     & $.720{\scriptstyle\,\pm.004}$ & $.735{\scriptstyle\,\pm.002}$ & $.324{\scriptstyle\,\pm.007}$ & $.575{\scriptstyle\,\pm.007}$ & $.588{\scriptstyle\,\pm.004}$ \\
    \bottomrule
  \end{tabular}}
\end{table}

\subsection{Single-axis baselines and step matching}
\label{sec:appendix-single-axis}

To verify the advantage of joint four-axis adaptation over single-axis specialization directly, we train a temporal-only baseline with the identical recipe, data pipeline, and budget as Axis-LoRA, differing only in using the TEMPORAL counterfactual triplets and no routing (temporal is the most under-exposed axis in our diagnosis, so the strongest single-axis choice). Because the temporal subset is smaller ($3{,}220$ of $4{,}196$ triplets), we additionally train a \emph{step-matched} variant for the same number of optimizer steps as the joint model.

\begin{table}[t]
  \centering
  \footnotesize
  \setlength{\tabcolsep}{5pt}
  \caption{Single-axis baselines, $\Delta$ 4-benchmark mean R@10 (pp) over Base, $4$-seed means.}
  \label{tab:single-axis}
  \resizebox{\columnwidth}{!}{%
  \begin{tabular}{@{}lccc@{}}
    \toprule
    & MiniLM & bge-base & Qwen3-0.6B \\
    \midrule
    Temporal-only                  & $+0.56$ & $+0.38$ & $-0.61$ \\
    Temporal-only (step-matched)   & $+0.92$ & $+0.77$ & $-0.94$ \\
    \textbf{Joint 4-axis Axis-LoRA} & $\mathbf{+1.56}$ & $\mathbf{+1.38}$ & $\mathbf{+1.28}$ \\
    \bottomrule
  \end{tabular}}
\end{table}

The temporal-only baseline recovers at most a third of the joint gain on MiniLM and bge-base and falls below Base on Qwen3-0.6B (Table~\ref{tab:single-axis}); matching the optimizer steps narrows but does not close the gap, and on Qwen3-0.6B the additional temporal-only training deepens the deficit. Nor is Qwen3's negative number an artifact of picking a weak axis: training each of the four single-axis baselines for Qwen3-0.6B under the same recipe gives $-0.61$ (TEMPORAL), $+0.22$ (SPATIAL), $-0.93$ (ENTITY), and $-0.97$ (EVENT), so even the best single axis recovers under a fifth of the joint gain. The most useful axis differs by encoder, and joint four-axis training is what makes the gain positive and consistent across encoders.

\subsection{Shared-trunk: capacity matching and training stability}
\label{sec:appendix-shared-trunk}

Two controls confirm that the shared-trunk collapse of Table~\ref{tab:3way-ablation} reflects the design rather than optimization instability. First, a capacity-matched shared trunk with rank $64$, matching the total adapter parameter count of the four rank-$16$ branches, still collapses under the identical recipe ($0.615{\scriptstyle\,\pm.021}$ / $0.572{\scriptstyle\,\pm.006}$ / $0.248{\scriptstyle\,\pm.018}$ / $0.428{\scriptstyle\,\pm.008}$ on LME / ATM-N / ATM-H / LoCoMo, 4-bench mean $0.465{\scriptstyle\,\pm.009}$, below the rank-$16$ trunk's $0.493$), so added capacity does not recover performance. Second, training is stable: the smoothed contrastive loss descends steadily to a plateau and never diverges (Table~\ref{tab:shared-trunk-loss}).

\begin{table}[t]
  \centering
  \footnotesize
  \setlength{\tabcolsep}{5pt}
  \caption{Shared-trunk contrastive training loss ($4$-seed mean) at $0$--$100\%$ of training steps. Both ranks converge without divergence.}
  \label{tab:shared-trunk-loss}
  \resizebox{\columnwidth}{!}{%
  \begin{tabular}{@{}lcccccc@{}}
    \toprule
    & $0\%$ & $20\%$ & $40\%$ & $60\%$ & $80\%$ & $100\%$ \\
    \midrule
    Shared trunk, $r{=}16$ & $0.90$ & $0.75$ & $0.65$ & $0.60$ & $0.51$ & $0.45$ \\
    Shared trunk, $r{=}64$ & $0.77$ & $0.64$ & $0.50$ & $0.58$ & $0.52$ & $0.55$ \\
    \bottomrule
  \end{tabular}}
\end{table}

The loss converges without divergence, the results are reproducible across seeds (stds $\leq 0.021$ per cell), and quadrupling capacity does not help. The collapse is a stable property of forcing four axis-decomposed outputs through shared weights.

\section{Experimental setup}
\label{sec:appendix-experimental-setup}

\subsection{Evaluation benchmark details}
\label{sec:appendix-eval-data}

Table~\ref{tab:bench-stats} summarizes per-split statistics for the four evaluation splits referenced in \S\ref{sec:results-setup}.

\begin{table}[h]
  \centering
  \footnotesize
  \setlength{\tabcolsep}{4pt}
  \caption{Per-split statistics. \emph{Pool} is the memory pool the retriever ranks over. \emph{Avg evid.} is the mean number of gold supporting memories per query; \emph{Multi-evid.} is the fraction of queries with more than one gold memory.}
  \label{tab:bench-stats}
  \resizebox{\columnwidth}{!}{%
  \begin{tabular}{@{}llrrcc@{}}
    \toprule
    Benchmark & Split & Queries & Pool & Avg evid. & Multi-evid. \\
    \midrule
    LongMemEval & ---    & $500$       & $940$        & $1.90$ & $65\%$ \\
    LoCoMo      & ---    & $1{,}977$   & $5{,}882$    & $1.42$ & $21\%$ \\
    ATM-Bench   & Normal & $1{,}013$   & $11{,}034$   & $1.44$ & $25\%$ \\
    ATM-Bench   & Hard   & $31$        & $11{,}034$   & $6.26$ & $100\%$ \\
    \bottomrule
  \end{tabular}}
\end{table}

\paragraph{Sample queries.} One representative query per split:
\begin{itemize}
  \setlength{\itemsep}{2pt}
  \item \textbf{LongMemEval}: \emph{``[2023/04/10] What was the first issue I had with my new car after its first service?''}
  \item \textbf{LoCoMo}: \emph{``When did Caroline go to the LGBTQ support group?''}
  \item \textbf{ATM-Bench Normal}: \emph{``Which hotel did I stay in on the first day of my trip to Kent?''}
  \item \textbf{ATM-Bench Hard}: \emph{``I really like the steel bridge in Porto. Can you help me find all the images and videos I took during my visit?''}
\end{itemize}

\subsection{Per-model training hyperparameters}
\label{sec:appendix-hp-table}

\paragraph{Shared method hyperparameters.} The following hyperparameters are identical across all three embedding models: LoRA rank $r=16$ applied to the last $20\%$ of attention layers on \texttt{q\_proj} and \texttt{v\_proj} with dropout $0.05$; InfoNCE temperature $\tau=0.05$; gate temperature annealed $\tau_g\!:2.0 \!\to\! 0.5$; KD weight $\lambda_{\mathrm{KD}}=0.3$ over a $50$K MS~MARCO triplet pool; inference mixing weight $\alpha=0.25$; AdamW with adapter learning rate $2 \times 10^{-4}$, weight decay $10^{-4}$, and warmup ratio $0.05$.

\begin{table*}[t]
  \centering
  \footnotesize
  \setlength{\tabcolsep}{8pt}
  \caption{Per-architecture optimization hyperparameters. Shared method hyperparameters are listed in the surrounding text.}
  \label{tab:hp-per-model}
  \begin{tabular}{@{}lccc@{}}
    \toprule
    & MiniLM-L6-v2 & bge-base-en-v1.5 & Qwen3-Embedding-0.6B \\
    \midrule
    Max sequence length              & $512$              & $512$              & $1{,}536$ \\
    Per-GPU batch size               & $64$               & $16$               & $2$ \\
    Gradient accumulation            & $1$                & $1$                & $2$ \\
    Effective batch ($\times 4$ GPU) & $256$              & $64$               & $16$ \\
    Epochs                           & $100$              & $5$                & $5$ \\
    Learning rate (gate)             & $1 \times 10^{-3}$ & $5 \times 10^{-4}$ & $5 \times 10^{-4}$ \\
    MS~MARCO per batch               & $8$                & $8$                & $2$ \\
    \bottomrule
  \end{tabular}
\end{table*}

\paragraph{Selection process.} The inference mixing weight $\alpha$ is selected via a documented sweep on Qwen3-Embedding-0.6B (Appendix~\ref{sec:alpha-sweep}). LoRA rank ($r=16$) and depth (last $20\%$ of attention layers) were chosen by a small grid on a single seed of Qwen3-Embedding-0.6B; rank expansion from $r=8$ to $r=16$ added $+0.30$~pp 4-bench mean at $76\%$ more trainable parameters (still $0.26\%$ of base), and depth expansion was not explored due to GPU memory limits. The InfoNCE temperature and AdamW with linear warmup are standard defaults from dense-retrieval training. The gate annealing schedule, KD weight, and adapter learning rate were set by small grid sweeps on Qwen3-Embedding-0.6B during development (a $3{\times}3$ grid over the KD weight and the final gate temperature, plus learning-rate sweeps), selected by R@10 on the held-out AxisCF validation split. Per-model optimization hyperparameters (max sequence length, batch size, epochs) are chosen to fit each architecture's memory footprint and training dynamics. Everything else is selected on Qwen3-Embedding-0.6B and transferred to MiniLM-L6-v2 and bge-base-en-v1.5 without retuning.

\paragraph{Query instruction prefixes.} Each model is evaluated with its model-card-recommended query prefix: MiniLM-L6-v2 uses no prefix; bge-base-en-v1.5 uses ``Represent this sentence for searching relevant passages:\,''; Qwen3-Embedding-0.6B uses ``Instruct: Given a personal-memory question, retrieve relevant memory entries.$\backslash$nQuery:\,'', a personal-memory-targeted instantiation of Qwen3's task-instruction interface.

\paragraph{Axis-specific instructions (axis-prefix prompt baseline).} The Axis-prefix prompt variant in Table~\ref{tab:3way-ablation} replaces the default Qwen instruction with four axis-specific instructions: \textbf{TEMPORAL} ``Instruct: Given a question about WHEN (time, date, period) something happened, retrieve relevant memory entries.''; \textbf{SPATIAL} ``Instruct: Given a question about WHERE (location, place) something happened, retrieve relevant memory entries.''; \textbf{ENTITY} ``Instruct: Given a question about WHO or WHAT (people, objects) is involved, retrieve relevant memory entries.''; \textbf{EVENT} ``Instruct: Given a question about WHAT EVENT or ACTION occurred, retrieve relevant memory entries.''. All four are followed by ``$\backslash$nQuery:\,''.

\paragraph{Software environment.} All reported evaluations were produced with a single pipeline under PyTorch 2.10.0 (CUDA 12.8), Transformers 5.6.2, and PEFT 0.19.1.

\subsection{MTEB BEIR-6 reproduction setup}
\label{sec:mteb-setup}

Table~\ref{tab:mteb-beir} reports per-task nDCG@10 for Qwen3-Embedding-0.6B base ($\alpha=0$) and + Ours ($\alpha=0.25$) on the BEIR-6 subset (NFCorpus, SciFact, ArguAna, SCIDOCS, FiQA2018, TRECCOVID). Both rows use the same evaluation configuration: max-len-q $256$, max-len-m $512$, and the general-retrieval query instruction prefix (``Instruct: Given a question, retrieve relevant passages that answer the question$\backslash$nQuery: ''). Our absolute base scores (mean $0.5449$) are below the official MTEB leaderboard report for Qwen3-Embedding-0.6B (mean $0.5882$): the gap is dominated by max-length truncation on the long-text tasks, since the official run uses task-specific instructions and a longer max-length budget. The relative $\Delta$ between the two rows is unaffected because both rows share the same shortened pipeline; this is the comparison we report in \S\ref{sec:results-main}.

\paragraph{Cross-encoder MTEB (MiniLM, bge-base).}
Table~\ref{tab:mteb-beir-cross} reports the same BEIR-6 sweep on the other two main-table encoders, each evaluated with its model-card-recommended query prefix (MiniLM: none; bge-base: ``Represent this sentence for searching relevant passages:'').
\begin{table}[t]
  \centering
  \footnotesize
  \setlength{\tabcolsep}{4pt}
  \caption{General retrieval (MTEB BEIR-6, nDCG@10) on MiniLM-L6-v2 and bge-base-en-v1.5. The two BERT-style embedding models show different preservation: bge-base near-perfect ($-0.02$~pp) and MiniLM mildly degraded ($-1.10$~pp).}
  \label{tab:mteb-beir-cross}
  \resizebox{\columnwidth}{!}{%
  \begin{tabular}{@{}lcccccc|cc@{}}
    \toprule
    \textbf{Setup} & NFC. & SciF. & ArgA. & SciD. & FiQA & TREC-C & \textbf{Avg.} & \textbf{$\Delta$} \\
    \midrule
    \multicolumn{9}{l}{\emph{MiniLM-L6-v2}} \\
    Base       & $0.3152$ & $0.6452$ & $0.4949$ & $0.2172$ & $0.3681$ & $0.4748$ & $0.4192$ & --- \\
    Axis-LoRA  & $0.3133$ & $0.6555$ & $0.4849$ & $0.2161$ & $0.3528$ & $0.4270$ & $0.4083$ & $-1.10$ \\
    \midrule
    \multicolumn{9}{l}{\emph{bge-base-en-v1.5}} \\
    Base       & $0.3773$ & $0.7443$ & $0.6026$ & $0.2219$ & $0.4002$ & $0.7672$ & $0.5189$ & --- \\
    Axis-LoRA  & $0.3747$ & $0.7501$ & $0.6216$ & $0.2174$ & $0.3951$ & $0.7538$ & $0.5188$ & $-0.02$ \\
    \bottomrule
  \end{tabular}}
\end{table}

The three embedding models show encoder-size-graded preservation: MiniLM (22M) shows the largest mean degradation ($-1.10$~pp), concentrated on TRECCOVID ($-4.78$~pp) and FiQA2018 ($-1.53$~pp); bge-base (110M) is near-perfect ($-0.02$~pp), with an ArguAna gain ($+1.90$~pp) partly offsetting modest drops on long-text splits; Qwen3-Embedding-0.6B (600M) sits in between at $-0.50$~pp (Table~\ref{tab:mteb-beir}), with the largest single drop on TRECCOVID ($-1.06$~pp). The smaller MiniLM has proportionally larger LoRA influence (last 20\% covers $6/12$ layers versus $5/28$ on Qwen), so the same axis-conditional reshaping translates to a larger perturbation of the general-retrieval geometry. The personalized-memory gains ($+1.28$ to $+1.56$~pp 4-bench mean R@10) more than offset the BEIR mean drops on every embedding model, but the trade-off is real and should be considered when transferring the recipe to small encoders.

\section{Additional results}
\label{sec:appendix-additional-results}

\subsection{Per-metric breakdown}
\label{sec:appendix-per-metric}

Table~\ref{tab:per-metric-breakdown} gives the full 4-seed per-metric breakdown (R@1, R@5, R@10, R@100, MRR) for the main 4-benchmark result of Table~\ref{tab:main-4bench}, on each (model, benchmark) cell.

\paragraph{Rank-redistribution mechanism.}
On Qwen3-Embedding-0.6B, R@5 dips slightly on 4-bench mean while R@10 lifts $+1.28$~pp; R@100 and MRR are essentially flat. Together these describe a rank-redistribution mechanism: the additive axis term reranks within the cosine-near band where \S\ref{sec:diagnosis-p1} identified residual axis signal that base cosine alone cannot rank. Rank-$11$ to $20$ axis-matching candidates move into the top-$10$. The pattern is sharpest on ATM-Hard ($+3.75$~pp R@10), consistent with multi-evidence queries benefiting most from per-axis ranking precision in the cosine-near band. On MiniLM-L6-v2 the gains are uniformly positive across all five aggregate metrics: the smallest model's baseline axis exposure is weak across all four axes (Table~\ref{tab:axis-scale-graded}), so Axis-LoRA lifts every retrieval band without rank redistribution. On bge-base-en-v1.5 the lift is concentrated on $3/4$ benchmarks (LME $+2.98$, ATM-Normal $+0.82$, ATM-Hard $+2.37$) with a small dip on LoCoMo ($-0.66$~pp), giving a 4-bench mean of $+1.38$~pp.

\begin{table*}[t]
  \centering
  \caption{Per-metric breakdown of the main 4-benchmark result ($4$-seed mean${\scriptstyle\pm}$std, $\alpha = 0.25$). $\Delta$ = Ours $-$ Base in pp.}
  \label{tab:per-metric-breakdown}
  \resizebox{\textwidth}{!}{%
  \begin{tabular}{@{}ll ccr ccr ccr ccr ccr@{}}
    \toprule
    & & \multicolumn{3}{c}{R@1} & \multicolumn{3}{c}{R@5} & \multicolumn{3}{c}{R@10} & \multicolumn{3}{c}{R@100} & \multicolumn{3}{c}{MRR} \\
    \cmidrule(lr){3-5} \cmidrule(lr){6-8} \cmidrule(lr){9-11} \cmidrule(lr){12-14} \cmidrule(lr){15-17}
    Encoder & Bench & Base & Ours & $\Delta$ & Base & Ours & $\Delta$ & Base & Ours & $\Delta$ & Base & Ours & $\Delta$ & Base & Ours & $\Delta$ \\
    \midrule
    \multirow{4}{*}{MiniLM}
    & LME    & $.198$ & $.203{\scriptstyle\pm.002}$ & $+0.4$ & $.378$ & $.398{\scriptstyle\pm.002}$ & $+2.1$ & $.474$ & $.487{\scriptstyle\pm.001}$ & $+1.3$ & $.816$ & $.819{\scriptstyle\pm.002}$ & $+0.3$ & $.382$ & $.392{\scriptstyle\pm.004}$ & $+1.1$ \\
    & ATM-N  & $.352$ & $.352{\scriptstyle\pm.002}$ & $+0.1$ & $.566$ & $.576{\scriptstyle\pm.003}$ & $+1.0$ & $.647$ & $.657{\scriptstyle\pm.002}$ & $+1.1$ & $.849$ & $.854{\scriptstyle\pm.001}$ & $+0.5$ & $.501$ & $.507{\scriptstyle\pm.001}$ & $+0.5$ \\
    & ATM-H  & $.037$ & $.059{\scriptstyle\pm.002}$ & $+2.2$ & $.224$ & $.238{\scriptstyle\pm.003}$ & $+1.4$ & $.281$ & $.305{\scriptstyle\pm.003}$ & $+2.5$ & $.633$ & $.659{\scriptstyle\pm.004}$ & $+2.5$ & $.328$ & $.396{\scriptstyle\pm.010}$ & $+6.8$ \\
    & LoCoMo & $.154$ & $.157{\scriptstyle\pm.001}$ & $+0.4$ & $.350$ & $.360{\scriptstyle\pm.002}$ & $+1.0$ & $.442$ & $.456{\scriptstyle\pm.002}$ & $+1.4$ & $.772$ & $.777{\scriptstyle\pm.002}$ & $+0.5$ & $.284$ & $.289{\scriptstyle\pm.001}$ & $+0.5$ \\
    \midrule
    \multirow{4}{*}{bge-base}
    & LME    & $.233$ & $.254{\scriptstyle\pm.003}$ & $+2.1$ & $.481$ & $.519{\scriptstyle\pm.003}$ & $+3.8$ & $.610$ & $.640{\scriptstyle\pm.003}$ & $+3.0$ & $.877$ & $.895{\scriptstyle\pm.000}$ & $+1.8$ & $.474$ & $.505{\scriptstyle\pm.002}$ & $+3.1$ \\
    & ATM-N  & $.435$ & $.437{\scriptstyle\pm.002}$ & $+0.2$ & $.663$ & $.664{\scriptstyle\pm.002}$ & $+0.1$ & $.729$ & $.737{\scriptstyle\pm.000}$ & $+0.8$ & $.898$ & $.898{\scriptstyle\pm.002}$ & $+0.0$ & $.597$ & $.601{\scriptstyle\pm.001}$ & $+0.4$ \\
    & ATM-H  & $.069$ & $.065{\scriptstyle\pm.003}$ & $-0.3$ & $.237$ & $.221{\scriptstyle\pm.003}$ & $-1.6$ & $.291$ & $.315{\scriptstyle\pm.008}$ & $+2.4$ & $.650$ & $.669{\scriptstyle\pm.007}$ & $+1.9$ & $.394$ & $.407{\scriptstyle\pm.012}$ & $+1.3$ \\
    & LoCoMo & $.277$ & $.268{\scriptstyle\pm.002}$ & $-0.9$ & $.518$ & $.517{\scriptstyle\pm.002}$ & $-0.1$ & $.624$ & $.618{\scriptstyle\pm.002}$ & $-0.7$ & $.878$ & $.880{\scriptstyle\pm.001}$ & $+0.2$ & $.436$ & $.428{\scriptstyle\pm.001}$ & $-0.8$ \\
    \midrule
    \multirow{4}{*}{Qwen3-0.6B}
    & LME    & $.291$ & $.298{\scriptstyle\pm.004}$ & $+0.7$ & $.617$ & $.616{\scriptstyle\pm.005}$ & $-0.1$ & $.717$ & $.724{\scriptstyle\pm.002}$ & $+0.6$ & $.940$ & $.940{\scriptstyle\pm.001}$ & $-0.0$ & $.579$ & $.588{\scriptstyle\pm.004}$ & $+0.9$ \\
    & ATM-N  & $.435$ & $.435{\scriptstyle\pm.002}$ & $+0.0$ & $.679$ & $.666{\scriptstyle\pm.005}$ & $-1.3$ & $.754$ & $.742{\scriptstyle\pm.003}$ & $-1.2$ & $.906$ & $.905{\scriptstyle\pm.003}$ & $-0.1$ & $.600$ & $.593{\scriptstyle\pm.002}$ & $-0.7$ \\
    & ATM-H  & $.075$ & $.056{\scriptstyle\pm.008}$ & $-1.9$ & $.228$ & $.188{\scriptstyle\pm.007}$ & $-3.9$ & $.299$ & $.337{\scriptstyle\pm.017}$ & $+3.8$ & $.747$ & $.762{\scriptstyle\pm.006}$ & $+1.5$ & $.416$ & $.383{\scriptstyle\pm.017}$ & $-3.3$ \\
    & LoCoMo & $.273$ & $.280{\scriptstyle\pm.003}$ & $+0.6$ & $.508$ & $.519{\scriptstyle\pm.002}$ & $+1.1$ & $.586$ & $.605{\scriptstyle\pm.003}$ & $+2.0$ & $.836$ & $.854{\scriptstyle\pm.001}$ & $+1.9$ & $.426$ & $.436{\scriptstyle\pm.002}$ & $+1.0$ \\
    \bottomrule
  \end{tabular}}
\end{table*}

\subsection{Bootstrap confidence intervals}
\label{sec:appendix-ci}

All intervals are query-level bootstraps: the per-query R@10 delta (Axis-LoRA $-$ Base), averaged over the $4$ seeds, is resampled $20$k times; 4-benchmark means use a hierarchical resample within each benchmark. Table~\ref{tab:ci-full} reports, alongside the 4-benchmark mean intervals quoted in \S\ref{sec:results-main}, the ATM-Hard split alone, the mean over the other three benchmarks, and the encoder-pooled ATM-Hard gain (per query, the gain is first averaged over the three encoders, then the $31$ queries are resampled jointly, preserving the correlation from evaluating all encoders on the same queries).

\begin{table}[t]
  \centering
  \footnotesize
  \setlength{\tabcolsep}{3pt}
  \caption{Query-level bootstrap $95\%$ CIs for R@10 gains (pp). At $n{=}31$ no single ATM-Hard interval excludes zero, reflecting the sample size; the gain is positive for all three encoders and significant when pooled across them.}
  \label{tab:ci-full}
  \resizebox{\columnwidth}{!}{%
  \begin{tabular}{@{}lcccccc@{}}
    \toprule
    & \multicolumn{2}{c}{ATM-Hard ($n{=}31$)} & \multicolumn{2}{c}{3-bench mean (excl.\ ATM-H)} & \multicolumn{2}{c}{4-bench mean} \\
    \cmidrule(lr){2-3} \cmidrule(lr){4-5} \cmidrule(lr){6-7}
    & $\Delta$ & 95\% CI & $\Delta$ & 95\% CI & $\Delta$ & 95\% CI \\
    \midrule
    MiniLM    & $+2.45$ & $[-0.85, +6.44]$ & $+1.26$ & $[+0.55, +1.99]$ & $+1.56$ & $[+0.54, +2.70]$ \\
    bge-base  & $+2.37$ & $[-1.42, +5.74]$ & $+1.05$ & $[+0.30, +1.82]$ & $+1.38$ & $[+0.25, +2.39]$ \\
    Qwen3-0.6B & $+3.75$ & $[-0.35, +8.48]$ & $+0.45$ & $[-0.16, +1.06]$ & $+1.28$ & $[+0.15, +2.52]$ \\
    \midrule
    3-encoder pooled & $\mathbf{+2.86}$ & $[\mathbf{+0.55}, +5.36]$ & & & & \\
    \bottomrule
  \end{tabular}}
\end{table}

\subsection{Per-query analysis of the regressing cells}
\label{sec:appendix-regression}

Two of the twelve (encoder, benchmark) cells regress (Qwen3-0.6B/ATM-Normal $-1.24$~pp, bge-base/LoCoMo $-0.66$~pp). Per-query ranks over the $4$ seeds (Table~\ref{tab:regression-perquery}) show the dips are a narrow rank-boundary effect rather than broad degradation: on both cells the large majority of queries keep the same top-$10$ outcome, and the net change comes from a small set of golds crossing the rank-$10$ cutoff, the rank-redistribution mechanism discussed with Table~\ref{tab:per-metric-breakdown}. In these two cells the exchange happens to be slightly unfavorable, while in the other ten it is favorable.

\begin{table}[t]
  \centering
  \footnotesize
  \setlength{\tabcolsep}{3pt}
  \caption{Per-query anatomy of the two regressing cells ($4$-seed averages). ``Unchanged'' = same top-$10$ hit outcome under Base and Axis-LoRA; ``wins/losses'' = golds that newly enter / drop out of the top-$10$ per seed; ranks are medians before $\to$ after.}
  \label{tab:regression-perquery}
  \resizebox{\columnwidth}{!}{%
  \begin{tabular}{@{}lcccc@{}}
    \toprule
    Cell & Unchanged & Wins / losses & Winner rank & Loser rank \\
    \midrule
    Qwen3 $\times$ ATM-N & $92\%$ ($933/1{,}013$) & $27$ / $50$ & $12 \to 7$ & $6 \to 12$ \\
    bge $\times$ LoCoMo  & $91\%$ ($1{,}794/1{,}977$) & $80$ / $103$ & $15 \to 6$ & $7 \to 12$ \\
    \bottomrule
  \end{tabular}}
\end{table}

\subsection{Query-type-stratified gains on real benchmarks}
\label{sec:appendix-stratified}

Using each benchmark's own question-type labels, only one type per dataset isolates a single axis, the TEMPORAL axis (\emph{temporal-reasoning} in LongMemEval, \emph{temporal} in LoCoMo); the other types blend several axes, so a gain there cannot be attributed to one axis. Across the six encoder$\times$benchmark cells for time questions, every encoder improves significantly on time questions in at least one of the two benchmarks, and no cell shows a significant loss (Table~\ref{tab:stratified-time}), matching the axis the diagnosis flags as most under-exposed. ATM-Bench labels by answer format rather than axis and cannot be split this way.

\begin{table}[t]
  \centering
  \footnotesize
  \setlength{\tabcolsep}{3pt}
  \caption{Time-question R@10 gains (pp) with query-level bootstrap $95\%$ CIs ($4$ seeds). Bold intervals exclude zero.}
  \label{tab:stratified-time}
  \resizebox{\columnwidth}{!}{%
  \begin{tabular}{@{}lcc@{}}
    \toprule
    & temporal-reasoning & temporal \\
    & (LME, $n{=}133$) & (LoCoMo, $n{=}320$) \\
    \midrule
    MiniLM     & $+1.32~[-2.4, +5.0]$          & $\mathbf{+2.99~[+0.8, +5.3]}$ \\
    bge-base   & $\mathbf{+4.46~[+1.1, +8.0]}$ & $-0.65~[-3.7, +2.2]$ \\
    Qwen3-0.6B & $+0.11~[-2.5, +2.7]$          & $\mathbf{+3.02~[+1.0, +5.2]}$ \\
    \bottomrule
  \end{tabular}}
\end{table}

\subsection{Axis-response replication on real benchmark memories}
\label{sec:appendix-real-memory}

We repeat the axis-response measurement of Table~\ref{tab:axis-scale-graded} on real memories drawn from the three evaluation benchmarks ($200$ ATM-Bench, $200$ LoCoMo, $100$ LongMemEval), building axis counterfactuals with the same generation pipeline and using the same 1-word content-edit floor. The pattern reproduces with the same axis ordering (Table~\ref{tab:real-memory}): TEMPORAL sits at or below the edit floor on all three encoders, the other axes clear the floor, and every axis stays at or below $11\%$ of the inter-memory distance, far below the scale at which cosine ranking discriminates. The pattern also holds within each source benchmark.

\begin{table}[t]
  \centering
  \footnotesize
  \setlength{\tabcolsep}{3pt}
  \caption{Axis-swap response on real benchmark memories: multiple of the 1-word edit floor / percentage of the inter-memory distance.}
  \label{tab:real-memory}
  \resizebox{\columnwidth}{!}{%
  \begin{tabular}{@{}lcccc@{}}
    \toprule
    & TEMPORAL & SPATIAL & ENTITY & EVENT \\
    \midrule
    MiniLM     & $0.05\times$ / $0.2\%$ & $1.41\times$ / $5.2\%$ & $2.57\times$ / $9.6\%$ & $1.64\times$ / $6.1\%$ \\
    bge-base   & $0.28\times$ / $1.1\%$ & $1.57\times$ / $6.3\%$ & $1.84\times$ / $7.4\%$ & $1.54\times$ / $6.2\%$ \\
    Qwen3-0.6B & $0.15\times$ / $0.8\%$ & $1.11\times$ / $6.0\%$ & $2.00\times$ / $10.8\%$ & $1.13\times$ / $6.1\%$ \\
    \bottomrule
  \end{tabular}}
\end{table}

\subsection{End-to-end QA under official protocols}
\label{sec:appendix-qa}

We evaluate how the retrieval gains carry through to answer quality, following each benchmark's official protocol: LongMemEval's reading prompt and per-question-type judge prompts, LoCoMo's QA prompt with normalized-token F1, and ATM-Bench's official evaluator. The top-$10$ memories from Base and from the seed-$42$ Axis-LoRA run are given to the same fixed reader (Qwen3.5-9B) and judge (Qwen3.5-27B) in both conditions, so the delta isolates retrieval. ATM-Hard is omitted because each of its $31$ questions carries a ${\sim}3$~pp weight on the split score, larger than the QA effects of interest. Retrieval on ATM-Hard, in contrast, is graded over ${\sim}6$ gold evidences per query and is reported with query-level confidence intervals (Appendix~\ref{sec:appendix-ci}).

\begin{table}[!htb]
  \centering
  \footnotesize
  \setlength{\tabcolsep}{5pt}
  \caption{End-to-end QA delta (Axis-LoRA $-$ Base, pp) under each benchmark's official metric (single seed).}
  \label{tab:qa-official}
  \resizebox{\columnwidth}{!}{%
  \begin{tabular}{@{}lccc@{}}
    \toprule
    & LME (judge acc.) & ATM-N (QS) & LoCoMo (F1) \\
    \midrule
    MiniLM     & $+2.80$ & $+0.60$ & $+0.90$ \\
    bge-base   & $+2.00$ & $-0.13$ & $-0.31$ \\
    Qwen3-0.6B & $+2.60$ & $-1.29$ & $+0.79$ \\
    \bottomrule
  \end{tabular}}
\end{table}

The QA delta (Table~\ref{tab:qa-official}) matches the sign of the retrieval delta in $8$ of $9$ cells, including the two retrieval regressions (Qwen3/ATM-Normal, bge/LoCoMo): retrieval gains and dips both propagate to answer quality. The coupling is also visible per query: on ATM-Normal, questions whose gold memory is retrieved in the top-$10$ are answered correctly $53.7\%$ of the time versus $5.0\%$ when it is missed. The magnitudes are compressed relative to the retrieval deltas because roughly $90\%$ of queries keep the same top-$10$ (Appendix~\ref{sec:appendix-regression}).

\end{document}